\documentclass[12pt]{article} %***
\usepackage[top=1.25in, bottom=1.25in, left=1in, right=1in]{geometry}
\usepackage[sectionbib]{natbib}
\usepackage{array,epsfig,fancyheadings,rotating,xcolor}
\usepackage[]{hyperref}  %<----modified by Ivan
\usepackage{sectsty, secdot}
\sectionfont{\fontsize{12}{14pt plus.8pt minus .6pt}\selectfont}
\renewcommand{\theequation}{\thesection\arabic{equation}}
\subsectionfont{\fontsize{12}{14pt plus.8pt minus .6pt}\selectfont}
\renewcommand{\baselinestretch}{2}
\usepackage{pdfpages}

\usepackage{amsmath}
\usepackage{amssymb}
\usepackage{amsfonts}
\usepackage{multirow}
\usepackage{amsthm}
\usepackage{bbm}
\usepackage{comment}
\usepackage{multirow}
\usepackage{rotating}
\usepackage{subfigure}
\usepackage{colortbl}
\usepackage{xr}
\usepackage{xr-hyper}

\newtheorem{theorem}{Theorem}

\newtheorem{proposition}{Proposition}
\theoremstyle{definition}

\newtheorem{remark}{Remark}
\usepackage{mathabx}

\newtheorem{condition}{Condition}
\newcommand{\Xc}{\mathcal{X}}
\newcommand{\Yc}{\mathcal{Y}}

\def\1v{\mathbf 1}
\def\0v{\mathbf 0}
\newcommand{\Ind}[1]{\mathbbm{1}{\left\{ {#1} \right\} }}

\newcommand{\R}{\mathbb R}

\newcommand{\var}{\mathop{\rm Var}}

\renewcommand{\thefootnote}{\fnsymbol{footnote}}

\def\spacingset#1{\renewcommand{\baselinestretch}%
	{#1}\small\normalsize} \spacingset{1}

\newcommand{\p}{\mathrm{pr}}

\begin{document}

%%%%%%%%%%%%%%%%%%%%%%%%%%%%%%%%%%%%%%%%%%%%%%%%%%%%%%%%%%%%%%%%%%%%%%%%%%%%%%%%%%%%%%%%%%%%%%%%%%%%%%%%%%%%%%%%%%%%%%%%%%%%
%%%%%%%%%%%%%%%%%%%%%%%%%%%%%%%%%%%%%%%%%%%%%%%%%%%%%%%%%%%%%%%%%%%%%%%%%%%%%%%%%%%%%%%%%%%%%%%%%%%%%%%%%%%%%%%%%%%%%%%%%%%%

\renewcommand{\baselinestretch}{2}

\markright{ \hbox{\footnotesize\rm %Statistica Sinica
%{\footnotesize\bf 24} (201?), 000-000
}\hfill\\[-13pt]
\hbox{\footnotesize\rm
%\href{http://dx.doi.org/10.5705/ss.20??.???}{doi:http://dx.doi.org/10.5705/ss.20??.???}
}\hfill }

\markboth{\hfill{\footnotesize\rm Zhao Ren, Sungkyu Jung and Xingye Qiao} \hfill}
{\hfill {\footnotesize\rm Covariance-engaged Classification of Sets via Linear Programming} \hfill}

\renewcommand{\thefootnote}{}
$\ $\par

%%%%%%%%%%%%%%%%%%%%%%%%%%%%%%%%%%%%%%%%%%%%%%%%%%%%%%%%%%%%%%%%%%%%%%%%%%%%%%%%%%%%%%%%%%%%%%%%%%%%%%%%%%%%%%%%%%%%%%%%%%%%

\fontsize{12}{14pt plus.8pt minus .6pt}\selectfont \vspace{0.8pc}
\centerline{\large\bf Covariance-engaged Classification of Sets via Linear Programming}
%\vspace{2pt} \centerline{\large\bf via Linear Programming}
\vspace{.4cm} \centerline{Zhao Ren$^{1}$, Sungkyu Jung$^{2}$ and Xingye Qiao$^{3}$} \vspace{.4cm} \centerline{\it
${}^{1}$University of Pittsburgh, ${}^{2}$Seoul National University, ${}^{3}$Binghamton University}
%\vspace{.2cm}
%\centerline{\it  ${}^{3}$Binghamton University}
\vspace{.55cm} \fontsize{9}{11.5pt plus.8pt minus
.6pt}\selectfont

%%%%%%%%%%%%%%%%%%%%%%%%%%%%%%%%%%%%%%%%%%%%%%%%%%%%%%%%%%%%%%%%%%%%%%%%%%%%%%%%%%%%%%%%%%%%%%%%%%%%%%%%%%%%%%%%%%%%%%%%%%%%

\begin{quotation}
\noindent {\it Abstract:}
Set classification aims to classify a set of observations as a whole, as opposed to classifying individual observations separately. To formally understand the unfamiliar concept of binary set classification, we first investigate the optimal decision rule under the normal distribution, which utilizes the empirical covariance of the set to be classified. We show that the number of observations in the set plays a critical role in bounding the Bayes risk. Under this framework, we further propose new methods of set classification. For the case where only a few parameters of the model drive the difference between two classes, we propose a  computationally-efficient approach to parameter estimation  using   linear programming, leading to the Covariance-engaged LInear Programming Set (CLIPS) classifier. Its theoretical properties are investigated for both independent case and various (short-range and long-range dependent) time series structures among observations within each set. The convergence rates of estimation errors and risk of the CLIPS classifier are established to show that having multiple observations in a set leads to faster convergence rates, compared to the standard classification situation in which there is only one observation in the set. The applicable domains in which the CLIPS performs better than competitors are highlighted in a comprehensive simulation study. Finally, we illustrate the usefulness of the proposed methods in classification of real image data in histopathology.

\vspace{9pt}
\par
\end{quotation}\par

\def\thefigure{\arabic{figure}}
\def\thetable{\arabic{table}}

\renewcommand{\theequation}{\thesection.\arabic{equation}}

\fontsize{12}{14pt plus.8pt minus .6pt}\selectfont

\setcounter{section}{0} %***
\setcounter{equation}{0} %-1

\lhead[\footnotesize\thepage\fancyplain{}\leftmark]{}\rhead[]{\fancyplain{}\rightmark\footnotesize\thepage}%Put this line in Page 2

%\noindent {\bf 1. Introduction}
\section{Introduction}
\label{sec:intro}
Classification is a useful tool in statistical learning with applications in many important fields. A classification method aims to train a classification rule based on the training data to classify future observations.  Some popular methods for classification include linear discriminant analyses, quadratic discriminant analyses, logistic regressions, support vector machines, neural nets and classification trees. Traditionally, the task at hand is to classify an observation into a class label. %training data comprise different individual observations, one for each object.

Advances in technology have eased the production of a large amount of data in various areas such as healthcare and manufacturing industries.  Oftentimes, multiple samples collected from the same object are available. For example,  it has become cheaper to obtain multiple tissue samples from a single patient in cancer prognosis \citep{miedema2012image}. To be explicit,  \cite{miedema2012image} collected 348 independent cells, each contains observations of varying numbers (tens to hundreds) of nuclei. Here, each cell, rather than each nucleus, is labelled as either normal or cancerous. Each observation of nuclei contains 51 measurements of shape and texture features.
A statistical task herein is to classify the whole set of observations from a single set (or all nuclei in a single cell) to normal or  cancerous group.
Such a problem was coined as \emph{set classification} by \cite{ning2009set}, studied in \cite{wang2012covariance} and \cite{Jung2014}, and was seen in the image-based pathology literature \citep{Samsudin2010,Wang2010,cheplygina2015classification,shifat2020cell} and in face recognition based on pictures obtained from multiple cameras, sometime called image set classification \citep{arandjelovic2006face,wang2012covariance}. The set classification is not identical to the multiple-instance learning (MIL) \citep{maron1998framework,chen2006miles,ali2010human,carbonneau2018multiple} as seen by \cite{kuncheva2010full}. A key difference is that in set classification a label is given to sets whereas  observations in a set have different labels in the MIL setting. % We defer a discussion on the connection between set classification and multiple-instance learning to Section \ref{sec:diss}.  %{\color{red}add comments on dependency after checking if literature has taken into account}

\begin{figure}[t!]
	\centering
	\includegraphics[width=0.5\textwidth]{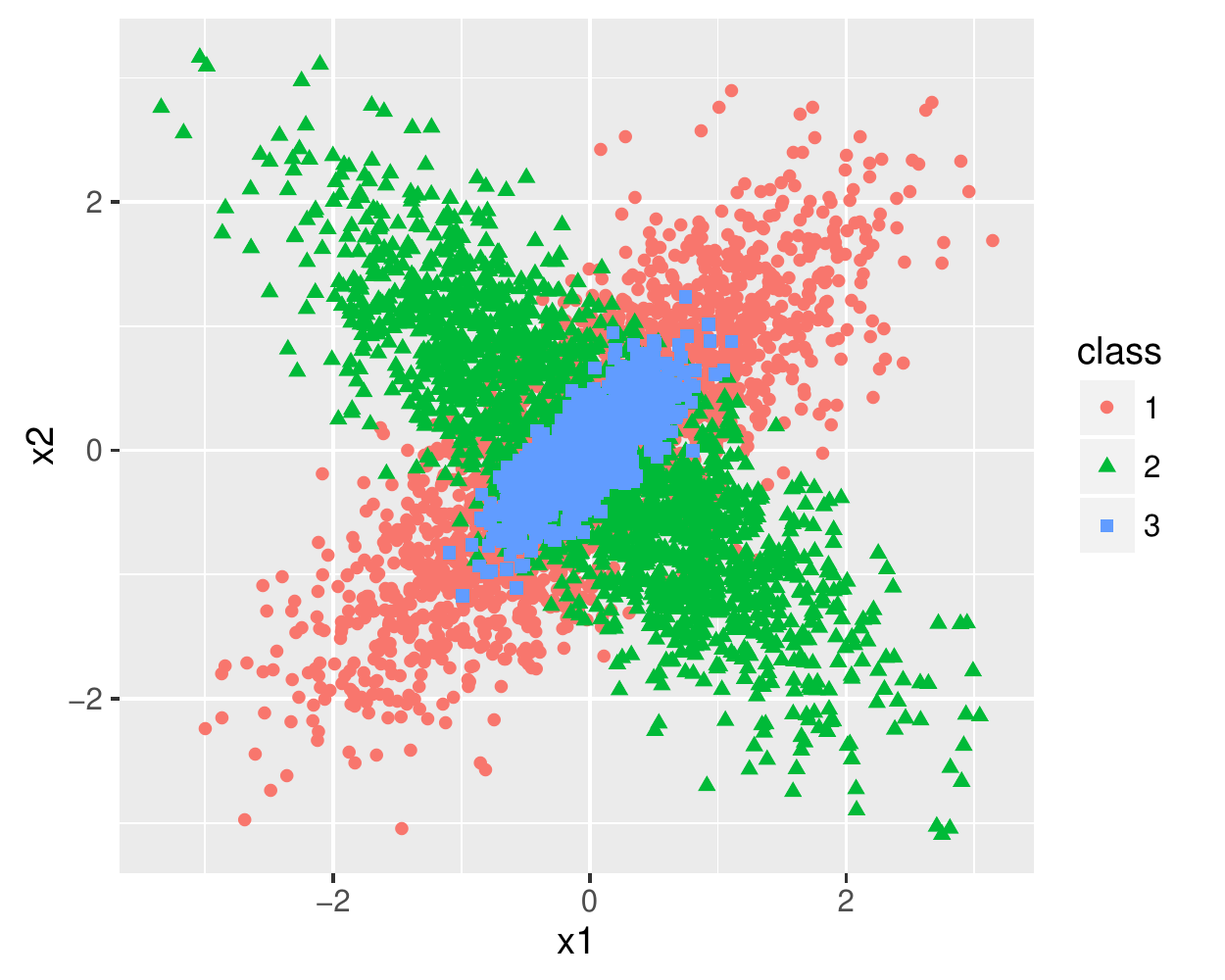}
	\caption{A 2-dimensional toy example showing classes with no difference in the mean or the marginal variance.}
	\label{fig:toy}
\end{figure}

While conventional classification methods predict a class label for each observation, care is needed in generalizing those for set classification. In principle, more observations should ease the task at hand. Moreover, higher-order statistics such as variances and covariances can now be exploited to help classification.
Our approach to set classification is to use the extra information, available to us only when there are multiple observations. To elucidate this idea, we illustrate samples from three classes in Fig.~\ref{fig:toy}. All three classes have the same mean, and Classes 1 and 2 have the same marginal variances.  Classifying a single observation near the mean to any of these distributions seems difficult. On the other hand, classifying several independent observations from the same class should be much easier. In particular, a set classification method needs to incorporate the difference in covariances to differentiate these classes.

In this work, we study a binary set classification framework, where a set of observations $\Xc = \{X_1,\ldots,X_M\}$ is classified to either $\Yc = 1$ or $\Yc = 2$. In particular, we propose set classifiers that extend quadratic discriminant analysis to the set classification setting, and are designed to work well in set-classification of high-dimensional data whose distributions are similar to those in Fig.~\ref{fig:toy}.

To provide a fundamental understanding of the set classification problem, we establish the Bayesian optimal decision rule under  normality and homogeneity (i.i.d) assumptions. This Bayes rule utilizes the covariance structure of the testing set of future observations. We show in Section~\ref{sec:setting} that it becomes much easier to make accurate classification for a set when the set size, $m_0$, increases. In particular, we demonstrate that the Bayes risk can be reduced exponentially in the set size $m_0$. To the best of our knowledge, this is the first formal theoretical framework for set classification problems in the literature.

Built upon the Bayesian optimal decision rule, we propose new methods of set classification in Section \ref{sec:methodologies}.
For the situation where the dimension $p$ of the feature vectors is much smaller than the total number of training samples, we demonstrate that a
simple plug-in classifier leads to satisfactory risk bounds similar to the Bayes risk. Again, a large set size   plays a key role in  significantly reducing the risk.
In high-dimensional situations where the number of parameters to be estimated ($\approx p^2$) is large, we make an assumption that only a few parameters drive the difference of two classes. With this sparsity assumption,  we propose to estimate the parameters in the classifier via linear programming, and the resulting classifiers are called Covariance-engaged LInear Programming Set (CLIPS) classifiers.
Specifically, the quadratic and linear parameters in the Bayes rule can be efficiently estimated under the sparse structure, thanks to the extra observations in the training set due to having sets of observations. %\textcolor[rgb]{0.00,0.00,1.00}{Therefore, in addition, our method is also adaptive in choosing between  linear and quadratic models.} %, and even elements of linear and quadratic parameters. }
Our estimation approaches are closely related to and built upon the successful estimation strategies in \cite{Cai2011constrained1minimization} and \cite{Cai2011direct}.
In estimation of the constant parameter, we perform a logistic regression with only one unknown, given the estimates of quadratic and linear parameters. This allows us to implement CLIPS classifier with high computation efficiency.

%
%we perform he constant parameter in the Bayes rule is estimated using a simple set-wise logistic regression.
%To gain a significant computational advantage, we perform a logistic regression with only one unknown  parameter, given the estimates of quadratic and linear parameters. %This strategy, however, imposes some challenges in theoretical analysis, due to the errors in the plugged-in estimates. %
%he proposed CLIPS classifier is computationally efficient as the estimation only involves linear programming and a simple logistic regression.

We provide a thorough study of theoretical properties of CLIPS classifiers and establish an oracle inequality in terms of the excess risk, in Section \ref{sec:theory quadratic}.
In particular, the estimates from CLIPS are shown to be consistent, and the strong signals are always selected with high probability in high dimensions. Moreover, the excess risk can be reduced by having more observations in a set, one of the new phenomena for set classification, which are different from that obtained by naively having pooled observations.

%We first demonstrate the advantages of having more observations under the homogeneity (i.i.d) assumption among all observations within each set. We then relax the independence assumption and investigate the behaviors of CLIPS classifiers under various (both short-range and long-range) dependent time series structures.  Results under both scenarios reveal new phenomena for set classification, which are different from that obtained by naively having pooled observations.

%with a factor $\sqrt{m_0}$ under some mild conditions, taking advantage of having $m_0$ observations in a set.

In the conventional classification problem where $m_0 = 1$, a special case of the proposed CLIPS classifier becomes a new sparse quadratic discriminant analysis (QDA) method \citep[cf.][]{fan2015innovated,fan2013optimal,Li2015Sparse,jiang2018direct,qin2018review,zou2019classification,gaynanova2019sparse,pan2020efficient}.
As a byproduct of our theoretical study, we show that the new QDA method enjoys better theoretical properties compared to  state-of-the-art sparse QDA methods such as   \cite{fan2015innovated}.

%In addition, we theoretically guarantee that our procedure chooses large signals in the quadratic parameters.

%is adaptive, in the sense of variable selection. For example,  automatically choose between quadratic and linear set classification.

%
%For example, when our procedure chooses quadratic parameter as an empty subset, then our CLIPS classifiers become a sparse LDA method
%In addition, our procedure  quadratic parameter %\ref{thm:quadratic}
%provides an empty subset, then our CLIPS classifiers become a sparse LDA method; see Theorem~\ref{thm:quadratic}.

%In high-dimensional settings, our proposed CLIPS classifiers are adaptive in the sense that they automatically choose between set classification and classical sparse QDA and sparse LDA.
%If the number of observations in each set $m_0=1$, then our CLIPS classifiers become a new sparse QDA procedure, which is related to many recently developed sparse QDA methods (\cite{fan2015innovated,fan2013optimal,Li2015Sparse}). The advantage of CLIPS classifiers is briefly discussed at the end of Section \ref{sec:theory quadratic}. In addition, if the subset selection result in Theorem \ref{thm:quadratic} provides an empty subset, then our CLIPS classifiers become a sparse LDA method.

%,fan2013optimal,Li2015Sparse}.

The advantages of our set classifiers are further demonstrated in comprehensive simulation studies. Moreover, we provide an application to histopathology in classifying sets of nucleus images to normal and cancerous tissues in Section \ref{sec:numerical}.
%Some further discussions, including the connection between set classification and multiple-instance learning as well as a robust version of CLIPS classifier against outliers are provided in Section \ref{sec:diss}.
Proofs of main results and technical lemmas can be found in the supplementary material. Also present in the supplementary material is a study on the case where observations in a set demonstrate certain spatial and temporal dependent structures. There, we utilize various (both short- and long-range) dependent time series structures within each set by considering a very general vector linear process model. %dependent time series structure to model the association among observations,

%%%%%%%%%%%%%%%%%%%%%%%%%%%%%%%%%%%%%%%%%
%%%%%%%%%%%%%%%%%%%%%%%%%%%%%%%%%%%%%%%%%

\section{Set Classification}\label{sec:setting}

%%%%%%%%%%%%%%%%%%%%%
%\subsection{Setting}\label{sec:setting}

We consider a binary set-classification problem. The training sample $\{(\mathcal{X}_{i},\mathcal{Y}_{i})\}_{i=1}^{N}$ contains $N$ sets of observations. Each set, $\mathcal{X}_{i}=\{X_{i1},X_{i2},\dots,X_{iM_{i}}\}\subset \mathbb{R}^p$, corresponds to one object, and is assumed to be from one of the two classes. The corresponding class label is denoted by  $\mathcal{Y}_{i}\in \{1,2\}$.
The number of observations within the $i$th set is denoted by $M_{i}$
and can be different among different sets. Given a new set of observations $(%
\mathcal{X}^{\dagger },\mathcal{Y}^{\dagger })$, the goal of {set
	classification}  is to predict $\mathcal{Y}^{\dagger }$  accurately based on $\mathcal{X}^{\dagger}$ using a classification rule $\phi (\cdot )\in \{1,2\}$ trained on the training sample.

To formally introduce set classification problem and study its fundamental properties, we start with a setting in which the sets in each class are homogeneous in the sense that all the observations in a class, regardless of the set
membership, follow the same distribution independently. Specifically, we assume both the $N$ sets $\{(\mathcal{X}_{i},\mathcal{Y}_{i})\}_{i=1}^{N}$ and the new set $(\mathcal{X}^{\dagger },\mathcal{Y}^{\dagger })$ are generated in the same way as $(\mathcal{X},\mathcal{Y})$ independently.
To describe the generating process of $(\mathcal{X},\mathcal{Y})$, we denote the marginal class probabilities by $\pi_1 = \p(\mathcal{Y}=1)$ and $\pi_2 = \p(\mathcal{Y}=2)$, and  the marginal distribution of the set size $M$ by $p_M$.
% and the marginal distribution of the set size $M\in \mathbb{N}$ is $p_{M}$.
We assume that the random variables $M$ and $\mathcal{Y}$ are independent. In other words, the class membership $\mathcal{Y}$ can not be predicted just based on the set size $M$. %the number of observations $M$ in a set does not imply whether the set is more likely to be from a class.
Conditioned on $M = m $ and $%
\mathcal{Y} = y $, observations $X_{1},X_{2},\dots ,X_{M}$ in the set $\Xc$ are independent and each distributed as $f_y$.

\subsection{Covariance-engaged Set Classifiers}\label{sec:method}

Suppose that there are $M^{\dagger}=m$ observations in the set $\mathcal{X}^{\dagger }=\{X_{1}^{\dagger },\dots
,X_{m}^{\dagger }\}$ that is to be
classified (called testing set),  and its true class label is $\mathcal{Y}%
^{\dagger }$.
The Bayes optimal decision rule classifies the set $\mathcal{X}^{\dagger }=\{x_{1},\dots ,x_{m}\}$ to Class 1
if the conditional class
probability of Class 1 is greater than that of Class 2, that is,
$\p(\mathcal{Y}^{\dagger }=1 \mid M^{\dagger }=m,~X_{j}^{\dagger
}=x_{j},~j=1,\dots,m) > 1/2$. This is equivalent to
$\pi _{1}p_M(m) \prod_{j=1}^m f_1(x_{j})>\pi _{2}p_M(m)\prod_{j=1}^m f_2(x_{j}),$
due to Bayes theorem and the independence assumption among $\mathcal{Y}^{\dagger }$ and $M^{\dagger }$.
%
%
%Denote the conditional class
%probability for Class 1 given $\mathcal{X}^{\dagger }$ as
%\begin{equation*}
%\eta (x_{1},\dots ,x_{m})=\p(\mathcal{Y}^{\dagger }=1\mid M^{\dagger }=m,~X_{i}^{\dagger
%}=x_{i},~i=1,\dots,m).
%\end{equation*}%
%The Bayes optimal decision rule classifies the set $\mathcal{X}^{\dagger }=\{x_{1},\dots ,x_{m}\}$ to class 1 if $\eta
%(x_{1},\dots ,x_{m})>0.5$.
%Due to the Bayes theorem and the independence assumption among $\mathcal{Y}^{\dagger }$ and $M^{\dagger }$,
%it is equivalent to that $\mathcal{X}^{\dagger }$ is classified to class 1 if
%Hence, the optimal decision is $1+\Ind{g(x_1,\dots,x_m)>0}$, where $g(x_1,\dots,x_m)$ is defined as $[\log\{f(x_1,\dots,x_m\mid Y=1)\}-\log\{f(x_1,\dots,x_m\mid Y=2)\}]/m$.
Let us now assume that the conditional distributions are both normal, that is, $%
f_1 \sim N(\mu_{1},\Sigma _{1})$ and $f_2 \sim N(\mu _{2},\Sigma _{2})$. Then the Bayes optimal decision rule depends on the quantity
\begin{align}
g(x_{1},\dots ,x_{m})  &=\frac{1}{m}\log \left\{\frac{\pi _{1}p_M(m)\prod_{j=1}^m f_1(x_{j})}{\pi _{2}p_M(m)\prod_{j=1}^m f_2(x_{j})}\right\} \notag \\
& =\frac{1}{m}\log (\pi _{1}/\pi _{2})-\frac{1}{2}\log (|\Sigma _{1}|/|\Sigma
_{2}|)-\frac{1}{2}\mu _{1}^{T}\Sigma _{1}^{-1}\mu _{1}+\frac{1}{2}\mu
_{2}^{T}\Sigma _{2}^{-1}\mu _{2}  \notag \\
& \qquad +(\Sigma _{1}^{-1}\mu _{1}-\Sigma _{2}^{-1}\mu _{2})^{T}\bar{x}+%
\frac{1}{2}\bar{x}^{T}(\Sigma _{2}^{-1}-\Sigma _{1}^{-1})\bar{x}+\frac{1}{2}%
\mbox{tr}\{(\Sigma _{2}^{-1}-\Sigma _{1}^{-1})S\}.  \label{eq:decision}
\end{align}%
Here $|\Sigma_k|$ denotes the determinant of the matrix $\Sigma_k$ for $k=1,2$, $\bar{x}=\sum_{j=1}^m x_j/m$ and $S=\sum_{j=1}^m(x_j-\bar{x})(x_j-\bar{x})^T/m$ are the sample mean and sample covariance of the testing set. Note that the realization $\mathcal{X}^{\dagger }=\{x_1,x_2,\dots,x_m\}$ implies both the number of observations $m$ and the i.i.d. observations $x_j$ for $j=1,\dots,m$. The Bayes rule can be expressed as
\begin{align}\label{eq:bayes2}
\phi _{B}(\mathcal{X}^{\dagger })
&= 2-\Ind{g(x_{1},\dots,x_{m})>0},\mbox{ where}\\
g(x_{1},\dots,x_{m})&= \frac{1}{m}\log (\pi _{1}/\pi_{2})+\beta_0+\beta^T\bar{x}+\bar{x}^{T}\nabla\bar{x}/2+\mbox{tr}(\nabla S)/2, \nonumber
\end{align}
in which the constant coefficient $\beta _{0}=\{-\log (|\Sigma _{1}|/|\Sigma
_{2}|)-\mu _{1}^{T}\Sigma _{1}^{-1}\mu _{1}+\mu _{2}^{T}\Sigma _{2}^{-1}\mu
_{2}\}/2\in \R$, the linear coefficient vector $\beta=\Sigma_{1}^{-1}\mu_{1}-\Sigma_{2}^{-1}\mu_{2}\in\R^p$ and the quadratic coefficient matrix $\nabla=\Sigma_{2}^{-1}-\Sigma_{1}^{-1}\in\R^{p\times p}$. The Bayes rule $\phi_B$ under the normal assumption in (\ref{eq:bayes2}) uses the summary statistics $m$, $\bar{x}$ and $S$ of $\mathcal{X}^{\dagger}$.

We refer to (\ref{eq:bayes2}) and any estimated version of it as a covariance-engaged set classifier. In Section \ref{sec:methodologies}, several estimation approaches for $\beta_0$, $\beta$ and $\nabla$ will be proposed. In this section, we further discuss a rationale for considering (\ref{eq:bayes2}).

The covariance-engaged set classifier (\ref{eq:bayes2}) resembles the conventional QDA classifier. As a natural alternative to (\ref{eq:bayes2}), one may consider the sample mean $\bar{x}$ as a representative of the testing set and apply QDA to $\bar{x}$ directly to make a prediction. In other words, one is about to classify this single observation $\bar{x}$ to one of the two normal distributions, that is, $f_1^{\prime} \sim N(\mu_{1},\Sigma _{1}/m)$ and $f_2^{\prime} \sim N(\mu _{2},\Sigma _{2}/m)$.
This simple idea leads to
\begin{align}\label{eq:tqda}
\phi _{B,\bar{x}}(\mathcal{X}^{\dagger })&=2-\Ind{ g_{\mathrm{QDA}}(\bar{x}) > 0 }, \mbox{ where}\\
g_{\mathrm{QDA}}(\bar{x})  &= \frac{1}{m}\log (\pi _{1}/\pi_{2})+\beta'_0+\beta^T\bar{x}+\bar{x}^{T}\nabla\bar{x}/2, \nonumber
\end{align}
in which  $\beta' _{0}=\{-\frac{1}{m}\log (|\Sigma _{1}|/|\Sigma
_{2}|)-\mu _{1}^{T}\Sigma _{1}^{-1}\mu _{1}+\mu _{2}^{T}\Sigma _{2}^{-1}\mu
_{2}\}/2$.
One major difference between (\ref{eq:bayes2}) and (\ref{eq:tqda}) is that the term $\mbox{tr}(\nabla S)/2$ is absent from (\ref{eq:tqda}). Indeed, the advantage of (\ref{eq:bayes2}) over (\ref{eq:tqda}) comes from the extra information in the sample covariance $S$ of $\mathcal{X}^{\dagger}$. In the regular classification setting, (\ref{eq:bayes2})  coincides with (\ref{eq:tqda})  since $\mbox{tr}(\nabla S)/2$ vanishes when $\mathcal{X}^{\dagger}$ is a singleton.

Given multiple observations in the testing set, another natural approach is  a majority vote applied to the QDA decisions of individual observations:
\begin{align}\label{eq:mv}
\phi _{MV}(\mathcal{X}^{\dagger })=2-\Ind{\frac{1}{m}\sum_{j=1}^m\mbox{sign}[  g_{\mathrm{QDA}}(x_j)  ] >0},
\end{align}
where $\mbox{sign}(t)=1,0,-1$ for $t>0,~t=0$ and $t<0$ respectively.
In contrast, since $g(\Xc^\dagger) = \frac{1}{m}\sum_{j=1}^mg_{\mathrm{QDA}}(x_j)$, our classifier (\ref{eq:bayes2}) predicts the class label by a weighted vote of individual QDA decisions. In this sense, the majority voting scheme (\ref{eq:mv}) can be viewed as a discretized version of (\ref{eq:bayes2}). In Section~\ref{sec:numerical}, we demonstrate that our set classifier (\ref{eq:bayes2}) performs significantly better than (\ref{eq:mv}).

\begin{remark}
	\label{remark:independence}
	We have assumed that $M$ and $\mathcal{Y}$ are independent in the setting. In fact, this assumption is not essential and can be relaxed. In a more general setting, there can be two different distributions of $M$, $p_{M1}(m)$ and $p_{M2}(m)$ conditional on $\mathcal{Y}=1$ and $\mathcal{Y}=2$ respectively. Our analysis {throughout the paper} remains the same except that they would replace two identical factors $p_M(m)$ in the first equality of (\ref{eq:decision}). If $p_{M1}(m)$ and $p_{M2}(m)$ are dramatically different, then the classification is easier as one can make decision based on the observed value of $m$. In this paper, we only consider the more difficult setting where $\mathcal{Y}$ and $M$ are independent.
\end{remark}

\subsection{Bayes Risk}\label{sec:bayesrisk}

%We now consider the theoretical aspect of the covariance-engaged set classifier, assuming that the parameter $\mu _{k}$ and $\Sigma _{k}$ for $k=1,2$ are known. In particular,

We show below an advantage of having a set of observations for prediction, compared to having a single observation. For this, we suppose for now that the parameters $\mu _{k}$ and $\Sigma _{k}$, $k=1,2$, are known and make the following assumptions. Denote $\lambda_{\max }(A)$ and $\lambda_{\min }(A)$ as the greatest and smallest eigenvalues of a symmetric matrix $A$.

\begin{condition}\label{cond1}
	The spectrum of $\Sigma _{k}$ is bounded below and above: there
	exists some universal constant $C_e>0$ such that $C_e^{-1}\leq \lambda _{\min
	}(\Sigma _{k})\leq \lambda _{\max }(\Sigma _{k})\leq C_e$ for $k=1,2$.
\end{condition}
\begin{condition}\label{cond2}
	The support of $p_{M}$ is bounded between $c_{m}m_0$ and $C_{m}m_0$, where $c_m$ and $C_m$ are universal constants and $m_0=\mathbb{E}(M)$. In other words, $p_M(a)=0$ for any integer $a<c_{m}m_0$ or $>C_{m}m_0$. The set size $m_0$ can be large or growing when a sequence of models are considered.
\end{condition}
\begin{condition}\label{cond3}
	The prior class probability is bounded away from $0$ and $1$:
	there exists a universal constant $0<C_{\pi }<1/2$ such that  $C_{\pi }\leq \pi _{1},\pi _{2}\leq 1-C_{\pi }$.
\end{condition}

We denote $R_{Bk}=\p(\phi_B(\mathcal{X}^{\dagger})\neq k\mid \mathcal{Y}^{\dagger}=k)$ as the risk of the Bayes classifier (\ref{eq:bayes2}) given $\mathcal{Y}^{\dagger}=k$. Let $\delta =\mu_{2}-\mu_{1}$. For a matrix $B\in\R^{p\times p}$, we denote $\|B\|_F=(\sum_{i=1}^p\sum_{j=1}^pB_{ij}^2)^{1/2}$ as its Frobenius norm, where $B_{ij}$ is its $ij$th element. For a vector $a\in\R^p$, we denote $\|a\|=(\sum_{i=1}^pa_i^2)^{1/2}$ as its $\ell_2$ norm. The quantity $D_{p}=( \| \nabla \|_{F}^{2}+\|\delta
\|^{2}) ^{1/2}$ plays an important role in deriving a convergence rate of the Bayes risk $R_{B}=\pi _{1}R_{B1}+\pi _{2}R_{B2}$. Although the Bayes risk does not have a closed form, we show that under mild assumptions, it converges to zero at a rate on the exponent.

\begin{theorem}\label{thm:bayesrisk}
	Suppose that Conditions \ref{cond1}-\ref{cond3} hold. If $D_{p}^{2}m_0$ is sufficiently large, then $R_{B}\leq 4\exp \left( -c^{\prime }m_0D_{p}^{2}\right) $ for some small constant $c^{\prime }>0$ depending on $C_e$, $c_m$ and $C_{\pi}$ only. In particular, as $D_{p}^{2}m_0\rightarrow \infty ,$ we have $R_{B}\rightarrow 0$.
\end{theorem}

The significance of having a set of observations is illustrated by this fundamental theorem. When $p_{M}(1)=1$, which implies $M^{\dagger}\equiv 1$ and $m_0=1$, Theorem \ref{thm:bayesrisk} provides a Bayes risk bound $R_{B}\leq 4\exp \left( -c^{\prime }D_{p}^{2}\right)$ for the theoretical QDA classifier in the regular classification setting. To guarantee a small Bayes risk for QDA, it is clear that $D_{p}^{2}$ must be sufficiently large. In comparison, for the set classification to be successful, we may allow $D_{p}^{2}$ to be very close to zero, as long as $m_{0}D_{p}^{2}$ is sufficiently large. The Bayes risk of $\phi_B$ can be reduced exponentially in $m_0$ because of the extra information from the set.

We have discussed an alternative classifier via using the sample mean $\bar{x}$ as a representative of the testing set, leading to $\phi_{B,\bar{x}}$ (\ref{eq:tqda}). The following proposition quantifies its risk, which has a slower rate than that of Bayes classifier $R_B$.

\begin{proposition}\label{pro:meanrisk}
	Suppose that Conditions \ref{cond1}-\ref{cond3} hold. Denote the risk of classifier  $\phi_{B,\bar{x}}$ in (\ref{eq:tqda}) as $R_{\bar{x}}$. Assume $\Vert \nabla \Vert _{F}^{2}+ m_{0}\Vert \delta
	\Vert ^{2}$ is sufficiently large. Then $R_{\bar{x}}\leq 4\exp \left( -c'(\Vert \nabla \Vert _{F}^{2}+ m_{0}\Vert \delta\Vert ^{2})\right)$ for some small constant $c'>0$ depending on $C_e$, $c_m$ and $C_{\pi}$ only. In addition, the rate on the exponent cannot be improved in general, i.e., $R_{\bar{x}}\geq \exp \left( -c''(\Vert \nabla \Vert _{F}^{2}+ m_{0}\Vert \delta\Vert ^{2})\right)$ for some small constant $c''>0$.
\end{proposition}

\begin{remark}
	Compared to the result in Theorem \ref%
	{thm:bayesrisk}, the above proposition implies that classifier  $\phi_{B,\bar{x}}$ needs a stronger assumption but has a slower rate of convergence when the mean difference $m_0\Vert \delta \Vert ^{2}$
	is dominated by the covariance difference $\Vert \nabla \Vert _{F}^{2}$. After
	all, this natural $\bar{x}$-based classification rule only relies on the
	first moment of the data set $\mathcal{X}^{\dagger }$ while the sufficient
	statistics, the first two moments, are fully used by the covariance-engaged classifier in (\ref{eq:bayes2}).
\end{remark}

%\textcolor{cyan}{XQ: Zhao, please rewrite the statement of the theorem above and add more remarks.}

%%%%%%%%%%%%%%%%%%%%%%%%%%%%%%%%%%%%%%%%%
\section{Methodologies}\label{sec:methodologies}
%%%%%%%%%%%%%%%%%%%%%

We now consider estimation procedures for $\phi_B$ based on   $N$ training sets $\{(\mathcal{X}%
_{i},\mathcal{Y}_{i})\}_{i=1}^{N}$.
In Section \ref{sec:lowdim}, we first consider a moderate-dimensional setting where $p\leq c_0m_{0}N$ with a sufficiently small constant $c_0>0$. In this case we apply a naive
plug-in approach using natural estimators of the parameters $\pi
_{k}$, $\mu _{k}$ and $\Sigma _{k}$. A direct estimation approach using linear programming, suitable for high-dimensional data, is introduced in Section \ref{sec:highdim}.
Hereafter, $p=p(N)$ and $m_{0}=m_{0}(N)$ are considered as functions of
$N$ as $N$ grows.

%%%%%%%%%%%%%%%%%%%%%%%%%%%%%%%%%%%%%%%%%

\subsection{Naive Estimation Approaches}\label{sec:lowdim}
%%%%%%%%%%%%%%%%%%%%%

The prior class probabilities $\pi _{1}$
and $\pi _{2}$ can be consistently
estimated by the class proportions in the training data, $\hat{\pi}%
_{1}=N_{1}/N$ and $\hat{\pi}_{2}=N_{2}/N$, where $N_{k}=\sum_{i=1}^{N}%
\Ind{\mathcal{Y}_i=k}$.
Let $n_{k}=\sum_{i=1}^{N}M_{i}\Ind{\mathcal{Y}_i=k}$ denote the total sample size
for Class $k = 1,2$.
The set membership is ignored at the
training stage, due to the homogeneity assumption. Note $n_{k},$
$n_{1}+n_{2}$ and $N_{k}$ are random while $N$ is deterministic.
One can obtain consistent estimators of $\mu _{k}$ and $\Sigma _{k}$ based on the training data and plug them
in (\ref{eq:bayes2}). It is natural to use  the maximum likelihood
estimators given $n_{k}$,
\begin{equation}
\hat{\mu}_{k}=\sum_{(i,j):\mathcal{Y}_{i}=k}X_{ij}/{n_{k}}\mbox{ and }\hat{\Sigma}%
_{k}=\sum_{(i,j):\mathcal{Y}_{i}=k}\{(X_{ij}-\hat{\mu}_{k})(X_{ij}-\hat{\mu}%
_{k})^{T}\}/{n_{k}}. \label{eq: sample mean variance}
\end{equation}

For classification of $\mathcal{X}^{\dagger }=\{X_{1}^{\dagger
},\ldots ,X_{M^{\dagger }}^{\dagger }\}$  with $M^{\dagger }=m$, $X_i^{\dagger}=x_i$, the set classifier (\ref{eq:bayes2}) is estimated by
\begin{equation}
\hat{\phi}(\mathcal{X}^{\dagger })=2-\Ind{\frac{1}{m}\log
	(\hat{\pi}_{1}/\hat{\pi}_{2})+\hat\beta_0+\hat\beta^T\bar{x}+\bar{x}^{T}%
	\hat\nabla\bar{x}/2+\mbox{tr}(\hat\nabla S)/2>0}\label{eq:qsc_plugin},
\end{equation}%
where $\hat{\beta}_{0}=-\frac{1}{2}\left\{\log (|\hat{\Sigma}_{1}|/|\hat{%
	\Sigma}_{2}|)-\hat{\mu}_{1}^{T}\hat{\Sigma}_{1}^{-1}\hat{\mu}_{1}+%
\hat{\mu}_{2}^{T}\hat{\Sigma}_{2}^{-1}\hat{\mu}_{2}\right\}$, $\hat{\beta}%
=\hat{\Sigma}_{1}^{-1}\hat{\mu}_{1}-\hat{\Sigma}_{2}^{-1}\hat{\mu}_{2}$ and $%
\hat{\nabla}=\hat{\Sigma}_{2}^{-1}-\hat{\Sigma}_{1}^{-1}$. In (\ref{eq:qsc_plugin}) we have assumed $p < n_k$ so that $\hat\Sigma_k$ is invertible.

The generalization error of set classifier (\ref{eq:qsc_plugin}) is $\hat{R}=\pi
_{1}\hat{R}_{1}+$ $\pi _{2}\hat{R}_{2}$ where $\hat{R}_{k}=\p(\hat{\phi}(\mathcal{%
	X}^{\dagger })\neq k\mid \mathcal{Y}^{\dagger }=k)$. The classifier itself depends on
the training data $\{(\mathcal{X}_{i},\mathcal{Y}_{i})\}_{i=1}^{N}$ and
hence is random. In the equation above, $\p$ is understood as the
conditional probability given the training data. Theorem \ref{thm:qsc} reveals a theoretical property of $%
\hat{R}$ in a moderate-dimensional setting which allows $p,N,m_{0}$ to grow jointly. This includes the traditional setting in which $p$ is fixed.

\begin{theorem}\label{thm:qsc}
	Suppose that Conditions \ref{cond1}-\ref{cond3}
	hold. For any fixed $L>0$, if $D_{p}^{2}m_{0} \geq C_{0}$ for some sufficiently large $C_{0}>0$ and $p\leq c_{0}Nm_{0}$, $p^{2}/(Nm_{0}D_{p}^{2})\leq c_{0}$, $\log{p} \leq c_{0}N$ for
	some sufficiently small constant $c_{0}>0$, then with probability at least $1-O(p^{-L})$
	we have $\hat{R}\leq 4\exp \left( -c^{\prime }m_{0}D_{p}^{2}\right) $
	for some small constant $c^{\prime }>0$ depending on $C_{\pi},c_m, L$ and $C_e$.
\end{theorem}

In Theorem \ref{thm:qsc}, large values of  $m_{0}$ not only relax the assumption on $D_{p}$ but also
reduce the Bayes risk exponentially in $m_{0}$ with high probability.
%\begin{remark}
A similar result for QDA, where $M_i=M^{\dagger}\equiv 1$ and $m_{0}=1$, was obtained in \cite{Li2015Sparse} under a stronger
assumption $p^{2}/(ND_{p}^{2})\rightarrow 0$.    %See, for example, \cite{Li2015Sparse} for details.

%although we are able to classify the set successfully under weaker assumption\ $%
%D_{p}^{2}n\rightarrow \infty $ rather than $D_{p}^{2}\rightarrow n$ in
%the regular classification setting, the same requirement $p^{2}/(ND_{p}^{2})%
%\rightarrow 0$ may imply a stronger sampling size assumption on $p$ since
%$D_{p}^{2}$ can possibly very small.
%\end{remark}

%%

For the high-dimensional data where $p=p(N)\gg Nm_{0}$ and hence $p>n_{k}$
with probability $1$ for $k=1,2$ by Condition \ref{cond2}, it is problematic to plug in the
estimators (\ref{eq: sample mean variance}) since
$\hat{\Sigma}_{k}$  is rank deficient with probability $1$.
A simple remedy is to use a diagonalized or enriched version of $\hat{\Sigma}_{k}$, defined by
$\hat{\Sigma}_{k(d)}=\mbox{diag}\{(\hat{\sigma}_{k,ii})_{i=1,\dots ,p}\}$ or
$\hat{\Sigma}_{k(e)}=\hat{\Sigma}%
_{k}+\delta I_{p}$,
where $\delta >0$ and $I_p$ is a $p\times p$ identity matrix.
Both $\hat{\Sigma}_{k(d)}$ and $\hat{\Sigma}_{k(e)}$ are
invertible. However, to our best knowledge, no
theoretical guarantee has been obtained without some structural assumptions.

%%%%%%%%%%%%%%%%%%%%%
%%%%%%%%%%%%%%%%%%%%%%%%%%%%%%%%%%%%%%%%%
\subsection{A Direct Approach via Linear Programming}\label{sec:highdim}

%%%%%%%%%%%%%%%%%%%%%

To have reasonable classification performance in high-dimensional data analysis, one usually has to take advantage of certain extra information of the data or model. There are often
cases where only a few elements in $\nabla =\Sigma _{2}^{-1}-\Sigma
_{1}^{-1} $ and $\beta =\Sigma _{1}^{-1}\mu _{1}-\Sigma _{2}^{-1}\mu
_{2}$ truly drive the difference between the two classes. A naive plug-in
method proposed in Section \ref{sec:lowdim} has ignored such potential structure of the data.
We assume that both $\nabla $ and $%
\beta $ are known to be sparse such that only a few elements of those are nonzero.
In light of this, the Bayes decision rule  (\ref{eq:bayes2}) implies
the dimension of the problem can be significantly reduced, which makes consistency possible
even in the high-dimensional setting.

We propose to directly estimate the quadratic term $\nabla $, the linear term $\beta $ and the constant $\beta _{0}$
coefficients respectively, taking advantage of the assumed sparsity.
As the estimates are efficiently calculated by linear programming, the resulting classifiers are called Covariance-engaged Linear Programming Set (CLIPS) classifiers.
%Theoretical properties and generalization errors of CLIPS classifier are postponed to Section \ref{sec:theory quadratic}.

We first deal with the estimation of the quadratic term $\nabla =\Sigma
_{2}^{-1}-\Sigma _{1}^{-1}$, which is the difference between the two precision
matrices. We use some key techniques developed in the literature of precision matrix estimation  \citep[cf.][]{meinshausen2006high,bickel2008regularized,friedman2008sparse,yuan2010high,Cai2011constrained1minimization,ren2015asymptotic}.
These methods estimate a single precision matrix with a
common assumption that the underlying true precision matrix is sparse in
some sense. For the estimation of the difference, we propose to use a two-step thresholded estimator.

As the first step, we adopt the CLIME estimator \citep{Cai2011constrained1minimization} to obtain initial estimators $\tilde{%
	\Omega}_{1}$ and $\tilde{\Omega}_{2}$ of the precision matrices $\Sigma
_{1}^{-1}$ and $\Sigma _{2}^{-1}$. Let $\Vert B\Vert _{1}=\sum_{i,j}\left\vert
B_{ij}\right\vert $ and $\Vert B\Vert _{\infty }=\max_{i,j}\left\vert
B_{ij}\right\vert $ be the vector $\ell _{1}$ norm and vector supnorm of a $%
p\times p$ matrix $B$ respectively. The CLIME estimators are defined as
\begin{align}
\tilde{\Omega}_{k} =\mathop{\rm argmin}_{\Omega \in \mathbb{R}^{p\times
		p}}\Vert \Omega \Vert _{1} \mbox{ subject
	to }\Vert \hat{\Sigma}_{k}\Omega -I\Vert _{\infty }<\lambda _{1,N},~k=1,2, \label{eq:simpleEstimation0}
\end{align}
for some $\lambda _{1,N}>0$.

Having obtained $\tilde{%
	\Omega}_{1}$ and $\tilde{\Omega}_{2}$, in the second step, we take a
thresholding procedure on their difference, followed by a symmetrization to obtain our final estimator
$\tilde{\nabla} =(\tilde{\nabla}_{ij})$ where
\begin{align}
\tilde{\nabla}%
_{ij}=\min \{\breve{\nabla}_{ij},\breve{\nabla}_{ji}\},\breve{\nabla}_{ij}=(%
\tilde{\Omega}_{2,ij}-\tilde{\Omega}_{1,ij})\Ind{\left\vert
	\tilde{\Omega}_{2,ij}-\tilde{\Omega}_{1,ij}\right\vert >\lambda
	_{1,N}^{\prime }},  \label{eq:simpleEstimation1}
\end{align}
%\begin{eqnarray}
%%\notag \\
%\tilde{\nabla} &=&(\tilde{\nabla}_{ij})\mbox{ where }\tilde{\nabla}%
%_{ij}=\min \{\breve{\nabla}_{ij},\breve{\nabla}_{ji}\},\breve{\nabla}_{ij}=(%
%\tilde{\Omega}_{2,ij}-\tilde{\Omega}_{1,ij})\Ind{\left\vert
%\tilde{\Omega}_{2,ij}-\tilde{\Omega}_{1,ij}\right\vert >\lambda
%_{1,N}^{\prime }},  \label{eq:simpleEstimation1}
%\end{eqnarray}
for some thresholding
level $\lambda _{1,N}^{\prime }>0$.

Although this thresholded CLIME difference
estimator is obtained by first individually estimating $\Sigma _{k}^{-1}$,
we emphasize that the estimation accuracy only depends on the sparsity of their difference $\nabla$ rather than the sparsity of either $\Sigma _{1}^{-1}$ or $\Sigma _{2}^{-1}$ under a
relatively mild bounded matrix $\ell _{1}$ norm condition.
We will show in Theorem \ref{thm:quadratic} in Section \ref{sec:theory quadratic} that if the true precision matrix difference $\nabla$ is negligible,  $\tilde\nabla = 0$ with high probability. %; see Theorem \ref{thm:quadratic} in Section \ref{sec:theory quadratic} for further details.
When $\tilde\nabla = 0$, our method described in (\ref{eq:qsc_sparse}) becomes a linear classifier adaptively. The computation of $\tilde{\nabla}$ (\ref{eq:simpleEstimation1}) is
fast, since the first step (CLIME) can be recast as a linear
program and the second step is a simple thresholding procedure.

\begin{remark}
	As an alternative, one can also consider a direct estimation of $\nabla$ that does not rely on individual estimates of $\Sigma _{k}^{-1}$.
	For
	example, by allowing some deviations from the identity $\Sigma _{1}\nabla \Sigma
	_{2}-\Sigma _{1}+\Sigma _{2}=0$, \cite%
	{Zhao2014Direct} proposed to  minimize the vector $\ell _{1}$ norm of ${\nabla}$.
	Specifically, they proposed $\tilde{\nabla}^{ZCL}\in
	\mathop{\rm
		argmin}_{B}\Vert B\Vert _{1}$ subject to $\Vert \hat{\Sigma}_{1}B\hat{\Sigma}%
	_{2}-\hat{\Sigma}_{1}+\hat{\Sigma}_{2}\Vert _{\infty }\leq \lambda
	_{1,n}^{\prime \prime }$, where $\lambda _{1,n}^{\prime \prime }$ is some
	thresholding level.
	This method, however, is computationally expensive (as it has $O(p^{2})$ number of linear constraints when casted to linear programming) and can only
	handle relatively small size of $p$. See also \cite{jiang2018direct}. We chose to use (\ref{eq:simpleEstimation1}) mainly because of fast computation.
\end{remark}

Next we consider the estimation of the linear coefficient vector $\beta
=\beta _{1}-\beta _{2}$, where $\beta _{k}=\Sigma _{k}^{-1}\mu _{k}$, $k=1,2$. In the literature of sparse QDA and sparse LDA, typical sparsity assumptions
are placed on $\mu _{1}-\mu _{2}$ and $\Sigma _{1}-\Sigma _{2}$ \citep[see][]
{Li2015Sparse} or placed on both $\beta _{1}$ and $\beta _{2}$ \citep[see, for
instance][]{Cai2011direct,fan2015innovated}. In the latter
case, $\beta $ is also sparse as it is the difference of two sparse vectors.
For the estimation of $\beta$, we propose a new method which directly imposes sparsity on $\beta $,
without specifying the sparsity for $\mu _{k}$, $\Sigma _{k}$ or $\beta _{k}$ except for some relatively mild conditions (see Theorem \ref{thm:linear} for details.)

The true parameter $\beta _{k}$ satisfies $\Sigma _{k}\beta _{k}-\mu _{k}=0$.
However, due to the rank-deficiency of $\hat{\Sigma}_{k}$, there are either none or infinitely many $%
\theta _{k}$'s that satisfy an empirical equation $\hat{\Sigma}_{k}\theta _{k}-\hat{\mu}_{k}=0$.
Here, $\hat{\mu}_{k}$ and $\hat{\Sigma}_{k}$ are defined in (\ref{eq: sample mean variance}). We
relax this constraint and seek a possibly non-sparse pair  $(\theta _{1}, \theta _{2})$
with the smallest $\ell _{1}$ norm difference. We estimate
the coefficients $\beta $ by $\tilde{\beta}=\tilde{\beta}_{1}-\tilde{\beta}%
_{2}$, where
\begin{align}
(\tilde{\beta}_{1},\tilde{\beta}_{2})=\mathop{\rm argmin}_{(\theta _{1},\theta _{2}): \left\Vert \theta _{k}\right\Vert _{1}\leq L_{1}}\Vert \theta
_{1}-\theta _{2}\Vert _{1}\mbox{\rm  ~subject
	to }\Vert \hat{\Sigma}_{k}\theta _{k}-\hat{\mu}_{k}\Vert _{\infty }<\lambda
_{2,N},~k=1,2,\label{eq:simpleEstimation2}
\end{align}
where $L_{1}$ is some sufficiently large constant introduced only to ease theoretical evaluations.
In practice, the constraint $\left\Vert \theta
_{k}\right\Vert _{1}\leq L_{1}$ can be removed without affecting the
solution.
Note that \cite{jiang2018direct} proposed to estimate $(\Sigma_1^{-1} + \Sigma_2^{-1})(\mu_1 - \mu_2)$ rather than $\beta = \Sigma_1^{-1}\mu_1 - \Sigma_2^{-1}\mu_2$.
The direct estimation approach for $\beta $ above shares some
similarities with that of \cite{Cai2011direct}, especially in the relaxed $%
\ell _{\infty }$ constraint. However \cite{Cai2011direct} focused on a direct
estimation of $\Sigma ^{-1}(\mu _{2}-\mu _{1})$ for  linear discriminant
analysis in which $\Sigma=\Sigma_1=\Sigma_2$, while we target on $\Sigma _{2}^{-1}\mu _{2}-\Sigma
_{1}^{-1}\mu _{1}$ instead.
Our procedure (\ref%
{eq:simpleEstimation2}) can be recast as a linear programming problem \citep[see, for example,][]{candes2007dantzig,Cai2011direct} and is
computationally efficient.

Finally, we consider the estimation of the constant coefficient $\beta _{0}$%
.
%For the training data, recall we have $N$ sets of observations $\{(\mathcal{X}_{i},%
%\mathcal{Y}_{i})\}_{i=1}^{N}$ which have the same generating process. %The prior class probabilities are $\pi _{1}$
%and $\pi _{2}$ with $\pi _{1}+\pi _{2}=1$. %(When $\pi _{1}$ and $\pi
%_{2}$ are not available beforehand, they can be consistently estimated by
%using the class proportions in the training data, $\hat{\pi}_{1}=N_{1}/N$
%and $\hat{\pi}_{2}=N_{2}/N$. This may influence the rate of convergence
%though, Check the proof!!.)
The conditional class probability $\eta(x_1,\dots,x_m) = \p(\mathcal{Y}=1 \mid M=m,~X_{i}=x_{i},~i=1,\dots,m)$ that a set belongs to
Class $1$ given $\mathcal{X}=\{x_1, \ldots, x_m\}$ can be evaluated by the following logit
function,
\begin{align*}
\log \left\{ \frac{\eta(x_1,\dots,x_m)}{1-\eta(x_1,\dots,x_m)}\right\} =&
\log \frac{\pi _{1}}{\pi _{2}}+\log \left\{ \frac{\prod_{i=1}^mf_1(x_{i})}{\prod_{i=1}^mf_2(x_{i})}\right\}  \\
=& \log(\pi _{1}/\pi _{2})+m(\beta _{0}+\bar{x}^{T}\beta +\frac{1}{2}%
\bar{x}^{T}\nabla \bar{x}+\frac{1}{2}\mbox{tr}(\nabla S)),
\end{align*}%
where $\bar{x}$ and $S$ are the sample mean and covariance of the set $%
\{x_{1},\ldots ,x_{m}\}$ respectively. Having obtained our estimators $%
\tilde{\nabla}$ and $\tilde{\beta}$ from (\ref{eq:simpleEstimation1}) and (%
\ref{eq:simpleEstimation2}), and estimated $\hat{\pi}_{1}$ and $\hat{\pi}_{2}
$ by $N_{1}/N$ and $N_{2}/N$  from the training data, we have only a scalar $\beta _{0}$ undecided.
We may find an estimate $\tilde{\beta}_{0}$ by conducting a simple logistic
regression with dummy independent variable $M_i$ and offset $\log
(\hat{\pi}_1/\hat{\pi}_2)+M_i\left(\bar{X}_{i}^{T}\tilde{\beta}+\bar{X}_{i}^{T}\tilde{\nabla}\bar{X}_{i}/2+\mbox{\rm tr}(\tilde{\nabla}S_{i})/2\right)$
for the $i$th set of observations in the training data, where $M_i$, $\bar{X}_{i}$, and $S_{i}$ are sample size, sample mean, and sample covariance of the $i$th set. In particular, we solve
\begin{align}
\tilde{\beta}_{0}& =\mathop{\rm argmin}_{\theta _{0}\in \mathbb{R}}~\ell
(\theta _{0}\mid \{(\mathcal{X}_{i},\mathcal{Y}_{i})\}_{i=1}^{N},\tilde{\beta},%
\tilde{\nabla}),\mbox{\rm  where the negative log-likelihood is}
\label{eq:simpleEstimation3} \\
& \quad \ell (\theta _{0}\mid \{(\mathcal{X}_{i},\mathcal{Y}_{i})\}_{i=1}^{N},\tilde{\beta},\tilde{\nabla}) \label{eq:simpleEstimation4}\\
& =\frac{1}{N}\sum_{i=1}^{N}\Big((\mathcal{Y}_{i}-2)M_{i}\left( \theta _{0}+\frac{\log
	(\hat{\pi}_1/\hat{\pi}_2)}{M_{i}}+\bar{X}_{i}^{T}\tilde{\beta}+\bar{X}_{i}^{T}\tilde{%
	\nabla}\bar{X}_{i}/2+\mbox{\rm tr}(\tilde{\nabla}S_{i})/2\right)   \notag \\
& \quad  +\log \left[ 1+\exp \left\{ M_{i}\left( \theta _{0}+\frac{\log
	(\hat{\pi}_1/\hat{\pi}_2)}{M_{i}}+\bar{X}_{i}^{T}\tilde{\beta}+\bar{X}_{i}^{T}\tilde{%
	\nabla}\bar{X}_{i}/2+\mbox{\rm tr}(\tilde{\nabla}S_{i})/2\right) \right\} %
\right] \Big)  \notag
\end{align}%
Since there is only one independent variable in the logistic regression
above, the optimization can be easily and efficiently solved.

For the purpose of evaluating theoretical properties, we apply the sample splitting
technique~\citep{wasserman2009high,meinshausen2010stability}. Specifically, we randomly choose the first batch of $N_{1}/2$ and $N_{2}/2$ sets from two classes in the training data to obtain estimators $\tilde{%
	\nabla}$ and $\tilde{\beta}$ using (\ref{eq:simpleEstimation1}) and (\ref%
{eq:simpleEstimation2}). Then $\tilde{\beta}_{0}$ is estimated based on the second batch along with $\tilde{\nabla}$ and $\tilde{\beta}$ using (\ref%
{eq:simpleEstimation3}). We plug all the estimators in (\ref{eq:simpleEstimation1}), (\ref{eq:simpleEstimation2})
and (\ref{eq:simpleEstimation3}) into the Bayes decision rule (\ref{eq:bayes2})
and obtain the CLIPS classifier,
\begin{equation}
\tilde{\phi}(\mathcal{X}^{\dagger })=2-\Ind{\frac{\log
		(\hat{\pi}_1/\hat{\pi}_2)}{m}+\tilde\beta_0+\tilde\beta^T\bar{x}+\bar{x}^{T}\tilde\nabla%
	\bar{x}/2+\mbox{\rm tr}(\tilde\nabla S)/2>0},  \label{eq:qsc_sparse}
\end{equation}%
where $\bar{x}$ and $S$ are sample mean and covariance of $\mathcal{X}%
^{\dagger }$ and $M^{\dagger}=m$ is its size.

%\textcolor{red}{XQ: I propose an alternative here for your information (inspired by the melanoma data.)}
%\begin{align}
%\tilde{\beta}_{0}& =\mathop{\rm argmax}_{\theta _{0}\in \mathbb{R}}~\ell
%(\theta _{0}\mid (\mathcal{X},\mathcal{Y}),\tilde{\beta},\tilde{\nabla}),%
%\mbox{\rm  where the log-likelihood is}  \label{eq:simpleEstimation3} \\
%\ell (\theta _{0}\mid (\mathcal{X},\mathcal{Y}),\tilde{\beta},\tilde{\nabla}%
%)& =\sum_{i=1}^{N}\Big((2-Y_{i})\left\{ \theta _{0}+\log
%(N_{1}/N_{2})+\bar X_{i}^{T}\tilde{\beta}+\bar X_{i}^{T}\tilde{\nabla}\bar X_{i}/2+\textrm{tr}(\tilde{\nabla}S_i)/2\right\}
%\notag \\
%& \quad \quad -\log \left[ 1+\exp \left\{ \theta _{0}+\log
%(N_{1}/N_{2})+\bar X_{i}^{T}\tilde{\beta}+\bar X_{i}^{T}\tilde{\nabla}\bar X_{i}/2+\textrm{tr}(\tilde{\nabla}S_i)/2\right\} %
%\right] \Big)  \notag
%\end{align}%

%%%%%%%%%%%%%%%%%%%%%
\section{Theoretical Properties of CLIPS}
\label{sec:theory quadratic}
In this section, we derive the theoretical properties of the estimators from (\ref%
{eq:simpleEstimation1})--(\ref{eq:simpleEstimation3}) as well as
generalization errors for the CLIPS classifier (\ref{eq:qsc_sparse}). In particular, we demonstrate  the advantages of  having sets of independent observations in contrast to classical QDA setting with individual observations under the homogeneity assumption of Section \ref{sec:setting}. Parallel results under various time series structures can be found in the supplementary material.

%Section \ref{sec:theory indep} presents the advantages of  having sets of independent observations in contrast to classical QDA setting with individual observations.

% We further relax the independence assumption and allow various dependent time series structures in Section \ref{sec:theory dep} to fully demonstrate the properties of CLIPS classifier.

%\subsection{Theoretical Properties under the Independence Assumption}
\label{sec:theory indep}

To establish the statistical properties of the thresholded CLIME difference estimator $\tilde{%
	\nabla}$ defined in (\ref{eq:simpleEstimation1}), we assume that the true  quadratic parameter $\nabla
=\Sigma _{2}^{-1}-\Sigma _{1}^{-1}$ has no more than $s_{q}$ nonzero
entries,
\begin{equation}
\nabla \in \mathcal{FM}_{0}(s_{q})=\{A=(a_{ij})\in \mathbb{R}^{p\times p},%
\mbox{\rm
	symmetric}:\sum_{i,j=1}^{p}\Ind{a_{ij}\neq 0}\leq s_q\}.
\label{eq:matrix l_0 ball}
\end{equation}
Denote $\mathrm{supp(A)}$ as the support of the matrix $A$. We summarize the estimation error and a subset selection result in the following theorem.
\begin{theorem}
	\label{thm:quadratic} Suppose Conditions \ref{cond1}-\ref{cond3} hold.
	Moreover, assume $\nabla \in \mathcal{FM}_{0}(s_{q})$, $%
	\Vert \Sigma _{k}^{-1}\Vert _{\ell_{1}}\leq C_{\ell1}$ with some constant $%
	C_{\ell1}>0$ for $k=1,2$ and $\log p\leq c_0N$ with some sufficiently small constant $c_0>0$. Then for any fixed $L>0,$ with
	probability at least $1-O(p^{-L})$, we have that
	\begin{eqnarray*}
		\Vert \tilde{\nabla}-\nabla \Vert _{\infty } &\leq &2\lambda _{1,N}^{\prime
		}, \\
		\Vert \tilde{\nabla}-\nabla \Vert _{F} &\leq &2\sqrt{s_{q}}\lambda
		_{1,N}^{\prime },\\
		\Vert \tilde{\nabla}-\nabla \Vert _{1} &\leq &2s_{q}\lambda
		_{1,N}^{\prime },
	\end{eqnarray*}%
	as long as $\lambda _{1,N}\geq CC_{\ell1}\sqrt{\frac{\log p}{Nm_{0}}}$ and $%
	\lambda _{1,N}^{\prime }\geq 8C_{\ell1}\lambda _{1,N}$ in (\ref%
	{eq:simpleEstimation1}), where $C$ depends on $L,C_{e},C_{\pi }$ and $c_{m}$
	only. Moreover, we have $\p(\mathrm{supp(\tilde{\nabla})\subset supp(\nabla )})=1-O(p^{-L}).$
\end{theorem}

\begin{remark}
	The parameter space $\mathcal{FM}_{0}(s_q)$ can be easily extended into an
	entry-wise $\ell _{q}$ ball or weak $\ell _{q}$ ball with $0<q<1$ %
	\citep{abramovich2006special} and the estimation results in Theorem \ref%
	{thm:quadratic} remain valid with appropriate sparsity parameters. The subset selection result also remains true and the support of $\tilde{\nabla}$
	contains those important signals of $\nabla $ above the noise level $\sqrt{%
		(\log p)/Nm_{0}}$. To simplify the analysis, we only consider $\ell _{0}$
	balls in this work.
\end{remark}

\begin{remark}
	Theorem \ref{thm:quadratic} implies that both the error bounds of estimating $%
	\nabla $ under vector $\ell_1$ norm and Frobenius norm rely on the sparsity $s_{q}$
	imposed on $\nabla $ rather than those imposed on $\Sigma _{2}^{-1}$ or $%
	\Sigma _{1}^{-1}$. Therefore, even if both $\Sigma _{2}^{-1}$ and $\Sigma
	_{1}^{-1}$ are relatively dense,  we still have an accurate estimate of $\nabla $ as long as $\nabla $ is very sparse and $%
	C_{\ell1}$ is not large.
\end{remark}

The proof of Theorem \ref{thm:quadratic}, provided in the supplementary material, partially follows from  \cite{Cai2011constrained1minimization}.

Next we assume $\beta =\beta _{1}-\beta _{2}$ is
sparse in the sense that it belongs to the $s_{l}$-sparse ball,
\begin{equation}
\beta
\in\mathcal{F}_{0}(s_l)=\{\alpha =(a_{j})\in \mathbb{R}^{p}:\sum_{j=1}^{p}\Ind{%
	\alpha _{j}\neq 0}\leq s_l\}.  \label{eq: l_q ball}
\end{equation}%
Theorem \ref{thm:linear} gives the rates of convergence of the linear
coefficient estimator $\tilde{\beta}$ in (\ref{eq:simpleEstimation2}) under the $\ell _{1}$ and $\ell _{2}$
norms. Both depend on the sparsity of $\beta $ only
rather than that of $\beta _{1}$ or $\beta _{2}$.

\begin{theorem}\label{thm:linear}
	 Suppose Conditions \ref{cond1}-\ref{cond3} hold.
	Moreover, assume that $\beta \in \mathcal{F}_{0}(s_{l})$, $\log p\leq c_0N$, $\Vert
	\beta _{k}\Vert _{1}\leq C_{\beta}$ and $\left\Vert \mu _{k}\right\Vert \leq
	C_{\mu}$ with some constants $C_{\beta},C_{\mu}>0$ for $k=1,2$ and some sufficiently small constant $c_0>0$. Then for any
	fixed $L>0,$ with probability at least $1-O(p^{-L})$, we have that
	\begin{eqnarray*}
		\Vert \tilde{\beta}-\beta \Vert _{1} &\leq &C^{\prime \prime}C_{\ell1}s_{l}\lambda _{2,N}, \\
		\Vert \tilde{\beta}-\beta \Vert &\leq &C^{\prime \prime}C_{\ell1}\sqrt{s_{l}}\lambda_{2,N},
	\end{eqnarray*}%
	as long as $\lambda _{2,N}\geq C^{\prime }\sqrt{\frac{\log p}{Nm_{0}}}$ in (%
	\ref{eq:simpleEstimation2}), where $\max \{\Vert \Sigma
	_{1}^{-1}\Vert _{\ell_{1}},\Vert \Sigma _{2}^{-1}\Vert _{\ell_{1}}\}\leq C_{\ell1}$ and $%
	C^{\prime \prime},C^{\prime }$ depend on $L,C_{e}, c_{m}, C_{\pi },C_{\beta}$ and $C_{\mu}$ only.
\end{theorem}

\begin{remark}
	The parameter space $\mathcal{F}_{0}(s)$ can be easily extended into an $\ell
	_{q}$ ball or weak $\ell _{q}$ ball with $0<q<1$ as well and the results in
	Theorem \ref{thm:linear} remain valid with appropriate sparsity parameters.
	We only focus on $\mathcal{F}_{0}(s)$ in this paper to ease the analysis.
\end{remark}

Lastly, we derive the rate of convergence for estimating the constant
coefficient $\beta _{0}$. Since $\tilde{\beta}_{0}$ is obtained by
maximizing the log-likelihood function after plugging $\tilde{\beta}$ and $%
\tilde{\nabla}$ in (\ref{eq:simpleEstimation3}), the behavior of our
estimator $\tilde{\beta}_{0}$ critically depends on the accuracy for estimating
$\beta$ and $\nabla$. Theorem \ref{thm:intercept} provides the result for $\tilde \beta _{0}$ based on certain general initial estimators $\tilde{\beta}$ and $%
\tilde{\nabla}$ with the following mild condition.

\begin{condition} \label{cond4}
	The expectation of the conditional variance of class label $\mathcal{Y}$
	given $\mathcal{X}$ is bounded below,
	that is, $\mathbb{E}\left(\var(\mathcal{Y}\mid \mathcal{X})\right)>C_{\log }>0$, where $C_{\log }$ is some universal constant.
\end{condition}

\begin{theorem}
	\label{thm:intercept} Suppose Conditions \ref{cond1}-\ref{cond4} hold, $\log p\leq c_0N$ with some sufficiently small constant $c_0>0$ and
	$\left\Vert \mu _{k}\right\Vert \leq C_{\mu}$ with some constant $C_{\mu}>0$ for $k=1,2$. Besides, we have some initial estimators $\tilde{\beta}$, $\tilde{\nabla}$, $\hat{\pi}_1$ and $\hat{\pi}_2$ such that $m_{0}(1+\sqrt{(\log p)/m_0}) \Vert \tilde{\beta}%
	-\beta \Vert +m_{0}(1+(\log p)/m_0) \Vert \tilde{\nabla}-\nabla \Vert _{1}+\max_{k=1,2} \vert \pi_k-\hat{\pi}_k\vert \leq C_{p}$  for
	some sufficiently small constant $C_{p}>0$ with probability at least $%
	1-O(p^{-L})$. Then, with probability at least $1-O(p^{-L}),$ we
	have%
	\begin{equation*}
	\left\vert \tilde{\beta}_{0}-\beta _{0}\right\vert \leq C_{\delta }\left((1+\sqrt{\frac{\log p}{m_{0}}})\Vert \tilde{\beta}-\beta \Vert +(1+\frac{\log p}{m_{0}})\Vert \tilde{\nabla}-\nabla \Vert _{1}+\max_{k=1,2}\frac{|\pi _{k}-\hat{\pi}_{k}|}{m_{0}} + \sqrt{\frac{\log p}{Nm^2_0}}\right),
	\end{equation*}%
	where constant $C_{\delta }$
	depends on $L,C_{e},C_{\pi },C_{\log },C_{\mu},C_m$ and $c_{m}$.
\end{theorem}

\begin{remark}
	Condition \ref{cond4} is determined by our data generating process stated in
	Section 2.1. It is satisfied when the classification problem is non-trivial. For example, it is valid if $\p\{C'<\p(\mathcal{Y}=1\mid \mathcal{X})<1-C'\}>C$ with some constants $C$ and $C'\in(0,1)$. As a matter of fact, Condition \ref{cond4} is weaker than the typical assumption: $C_{\log }<\p(\mathcal{Y%
	}=1\mid \mathcal{X})<1-C_{\log }$ with probability 1 for $\mathcal{X}$, which is
	often seen in the literature of logistic regression. See, for example, \cite%
	{fan2013asymptotic} and \cite{fan2015innovated}.
\end{remark}

Theorems \ref{thm:quadratic}, \ref{thm:linear} and \ref{thm:intercept} demonstrate the estimation
accuracy for the quadratic, linear and constant coefficients in our CLIPS classifier (\ref{eq:qsc_sparse}) respectively. We conclude this section by establishing an oracle inequality for its
generalization error via providing a rate of convergence of the excess risk. To this
end, we define the generalization error of CLIPS classifier as $%
\tilde{R}_{}=\pi _{1}\tilde{R}_{1}+\pi _{2}\tilde{R}_{2}$, where $\tilde{R}_{k}=\p(\tilde{\phi}(\mathcal{X}^{\dagger })\neq k\mid \mathcal{Y}^{\dagger }=k)$ is the
probability that a new set observation from Class $k$ is misclassified by the CLIPS classifier $\tilde{\phi}(\mathcal{X}^{\dagger })$. Again $\p$ is the conditional probability given  the training data $\{(\mathcal{X}_{i},\mathcal{Y}%
_{i})\}_{i=1}^{N}$ which $\tilde{\phi}(\mathcal{X}^{\dagger })$ depends on.

We introduce some notation related to the Bayes decision rule in (\ref{eq:bayes2}). Recall that given $M^{\dagger }=m$, the Bayes decision rule $%
\phi _{B}(\mathcal{X}^{\dagger })$ solely
depends on the sign of the function $g(\mathcal{X}^{\dagger }) = \frac{1}{m}\log (\pi_{1}/\pi%
_{2})+\beta_{0}+\beta^{T}\bar{x}+\bar{x}^{T}\nabla%
\bar{x}/2+\mbox{tr}(\nabla S)/2$. We define by $%
F_{k,m}$ the conditional cumulative distribution function of the oracle statistic $g(\mathcal{X}^{\dagger })$ given that $M^\dagger = m$ and $\mathcal{Y}^{\dagger }=k$.
The upper bound of the first derivatives of $F_{1,m}$ and $F_{2,m}$ for
all possible $m$ near $0$ is denoted by $d_{N}$,%
\begin{equation*}
d_{N}=\max_{m\in \lbrack c_{m}m_{0},C_{m}m_{0}],~k=1,2}\left\{\sup_{t\in \lbrack
	-\delta _{0},\delta _{0}]}\left\vert F_{k,m}^{\prime }(t)\right\vert \right\},
\end{equation*}%
where $\delta _{0}$ is any sufficiently small constant. The value of $d_{N}$
is determined by the generating process and is usually small whenever the Bayes
rule performs reasonably well. According to Theorems \ref{thm:quadratic}, \ref{thm:linear} and \ref{thm:intercept}, with probability at
least $1-O(p^{-L})$, our estimators satisfy that
\begin{equation*}
\Xi_N:= (1+\sqrt{\frac{\log p}{m_{0}}})\Vert \tilde{\beta}-\beta \Vert +(1+\frac{\log p}{m_{0}})\Vert \tilde{\nabla}-\nabla \Vert _{1}+
\max_{k=1,2}\frac{\vert \hat{\pi}_k-\pi_k\vert}{m_0}+\left\vert \tilde{\beta}_{0}-\beta
_{0}\right\vert = O(\kappa_N ),
\end{equation*}%
where $\kappa_N :=(1+(\log p)/m_0)s_{q}\lambda _{1,N}^{\prime }+(1+\sqrt{(\log p)/m_0})C_{\ell1}\sqrt{s_{l}}\lambda _{2,N}+\sqrt{(\log p)/(Nm^2_0)}$.
It turns out the quantity $\kappa_N d_{N} $ is the key to obtain the oracle inequality. Condition \ref{cond5} below guarantees that the assumptions of Theorem \ref{thm:intercept} are satisfied with high probability in our settings.

\begin{condition}\label{cond5}
	Suppose $\kappa_Nm_0\leq c_0$ and $\kappa_N d_{N}\leq c_0$  with some sufficiently small constant $c_0>0$.
\end{condition}

Theorem \ref{thm:oracle} below reveals the oracle property of CLIPS classifier and provides a rate of convergence of the excess risk, that is, the generalization error of CLIPS classifier less the Bayes risk $R_B$ defined in Section \ref{sec:bayesrisk}.

\begin{theorem}
	\label{thm:oracle} Suppose that the assumptions of Theorems \ref%
	{thm:quadratic} and \ref{thm:linear} hold and that Conditions \ref{cond4}--\ref{cond5} also hold.
	Then with probability at least $1-O(p^{-L})$, we have the oracle
	inequality%
	\begin{equation*}
	\tilde{R}_{}\leq R_{B}+C_g(\kappa_N d_{N}+p^{-L}),
	\end{equation*}%
	where constant $C_g$ depends on $L,C_{e},C_{\pi },C_{\log },C_{\beta},C_m,c_m$ and $C_{\mu}$ only. In particular, we have $\tilde{R}_{}$ converges to the Bayes risk $R_{B}$ in probability as $N$ goes to infinity.
\end{theorem}

Theorem \ref{thm:oracle} implies that with high probability, the
generalization error of CLIPS classifier is close to the Bayes risk with
rate of convergence no slower than $\kappa _{N}d_{N}$. In particular, whenever
the the quantities $d_{N}$ and $C_{\ell1}$ are bounded by some universal constant,
the thresholding levels $\lambda _{1,N}^{\prime }=O(\sqrt{\log p/(m_{0}N)})$
and $\lambda _{2,N}=O(\sqrt{\log p/(m_{0}N)})$ yield the rate of
convergence $\kappa_{N}d_{N}$ in the order of
\begin{equation}\label{eq:discussion.oracle}
(1+\sqrt{(\log p)/m_0})\sqrt{\log p/(m_{0}N)}\sqrt{s_{l}}+(1+(\log p)/m_0)\sqrt{\log p/(m_{0}N)}s_q.
\end{equation}

The advantage of having large $m_0$ can be understood by investigating (\ref{eq:discussion.oracle}) as a function of $m_0$. Indeed, the leading term of (\ref{eq:discussion.oracle}) is
\begin{align*}
\frac{\log p}{m_0^{3/2}}\sqrt{\frac{\log p}{N}} s_q, \ &\mbox{ if } m_0 \le \log p \cdot \min\{1,\frac{s_q^2}{s_l}\}; \\
\frac{\sqrt{\log p}}{m_0}\sqrt{\frac{\log p}{N}} \sqrt{s_l}, \ &\mbox{ if } \log p\cdot\frac{s_q^2}{s_l}\le m_0 \le \log p; \\
\sqrt{ \frac{1 }{ m_0}}\sqrt{\frac{\log p}{N}} (\sqrt{s_l}+s_q), \ &\mbox{ if } \log p \le m_0.
\end{align*}
%$\log p / \sqrt{N} \sqrt{ s^2 / m}$ if $m \le \log p$;
%$\sqrt{\log p/N} \sqrt{ s^2 / m}$  if $\log p \le m \le s^2$;
%$\sqrt{\frac{\log p}{N}} $ if $s^2 \le m$.
To illustrate the decay rate, we assume $s_l\geq s_q^2$. Then as $m_0$ increases, the error decreases at the order of $m_0^{3/2}$ up to certain point $\log p\cdot\frac{s_q^2}{s_l}$, and then decreases at the order of $m_0$ up to another point $\log p$.
When $m$ is large enough so that $m_0 \ge \log p$, then the error decreases at the order  of $\sqrt{m_0}$. %there is an additional gain at the order of $\sqrt{\log p}$. %However, when $m_0$ is too large that it exceeds $s^2$, then the last term $\sqrt{(\log p)/N}$, resulting essentially from estimating prior class probabilities $\pi_1$ and $\pi_2$, dominates. This may be improved with additional assumptions on the parameters. A similar phase transition phenomenon is also observed for the case $s^2 /\log p = O(1)$.

To further emphasize the advantage of having sets of observations, we compare a general case $m_0 = m^*$ where $\log p \le m^* $ with
the special case that $m_0 = 1$, i.e., the regular QDA situation.  Then the quantity $\kappa_{N}$ with $m^*$ has
a faster decay rate with a factor of order between $\sqrt{m^* \log p }$ and $\sqrt{m^*  }\log p$ (depending on the relationship between $s_l$ and $s_q$) compared to the $m_0 = 1$ case, thanks to the extra observations within each set.

\begin{remark} \label{rem:sample size independent}
The above discussion reveals that in high-dimensional setting the benefit of the set-classification cannot be simply explained by having $N^*=Nm_0$ independent observations instead of having only $N$ individual observations as in the classical QDA setting. Indeed, if we have $N^*$ individual observations in the classical QDA setting, then the implied rate of convergence would be either $\log p\sqrt{\frac{\log p}{Nm_0}} s_q$ (if $\log p\cdot s_q^2\geq s_l$) or $\sqrt{\log p}\sqrt{\frac{\log p}{Nm_0}} \sqrt{s_l}$ (otherwise), which is slower than the one provided in equation (\ref{eq:discussion.oracle}).
\end{remark}

\begin{remark}
%In the end, i
It is worthwhile to point out that even in the special QDA situation where $m_{0}=1$, due to the sharper analysis, our result is still new and the established rate of convergence $(\log p)/N^{1/2}\sqrt{s_{l}}+(\log p)^{3/2}/N^{1/2}s_{q}$ in  Theorem \ref{thm:oracle} is at least as good as the one $(\log p)^{3/2}/N^{1/2}(s_{q}+s_{l})$ derived in
the oracle inequality of \cite{fan2015innovated} under similar assumptions. Whenever $s_l>s_q$, our rate is even faster with a factor of order $\sqrt{s_l\log p  }$ than that in \cite{fan2015innovated}.
\end{remark}

\begin{remark} \label{rem:sample size_TS}
	Results in this section, including Theorem \ref{thm:oracle}, demonstrate the full advantages of the set classification setting in contrast to the classical QDA setting. When multiple observations within each set have short-range dependence, the rates of convergence for estimating key parameters as well as the oracle inequality resemble the results under independent assumption. However, the results significantly change when there is a long-range dependence structure among multiple observations.
\end{remark}

\section{Numerical Studies}\label{sec:numerical}
%\subsection{Implementation}
%To select the optimal value of $\lambda_{1,n}$, we conduct a grid search from $(0,\tilde\lambda_1]$ where $\tilde\lambda_1 = \|\hat{\Sigma }_{2}-\hat{\Sigma }_{1}\|_{\infty}.$ For any $\lambda_{1,n}\ge \tilde\lambda_1$, the solution to (\ref{eq:simpleEstimation1}) is 0, which means that the covariance term does not contribute much to classification.
%
%
%We select the optimal value of $\lambda_{2,n}$ from a fine grid taking values within $(\tilde\lambda_2^L,\tilde\lambda_2^U]$. If $\lambda_{2,n}<\tilde\lambda_2^L$, then either one of the two constraints in optimization problem (\ref{eq:simpleEstimation2}), $\| \hat{\Sigma}_{k}\theta _{k}-\hat{\mu}_{k}\| _{\infty}<\lambda _{2,n},~k=1,2$, is infeasible. If $\lambda_{2,n}\ge \tilde\lambda_2^U$, then the solution is trivial, that is, the solution is 0. In this case, the difference in the mean of the two classes do not contribute much to classification. Here, the lower end $\tilde\lambda_2^L$ is the maximum of the solutions $\hat c_1$ and $\hat c_2$ of the following two optimization problems, that is, $\tilde\lambda_2^L=\max(\hat c_1,\hat c_2)$ where for $k=1$ or $2$,
%\begin{align*}
%\hat c_k=\argmin_{c,\theta} c,~\textrm{subject to } c\ge 0\textrm{ and }\| \hat{\Sigma}_{k}\theta-\hat{\mu}_{k}\| _{\infty}\le c.
%\end{align*}
%The upper end $\tilde\lambda_2^U$ is found by
%\begin{align*}
%\lambda_2^U=\argmin_{c,\theta} c,~\textrm{subject to }c\ge 0\textrm{, and } \| \hat{\Sigma}_{k}\theta-\hat{\mu}_{k}\| _{\infty}\le c,~k=1\textrm{ and }2.
%\end{align*}

%%%%%%%%%%%%%%%%%%%%%%
In this section we compare various versions of covariance-engaged set classifiers with other set classifiers adapted from traditional methods. In addition to the CLIPS classifier, we use the diagonalized and enriched versions of $\hat\Sigma_k$ respectively (labeled as Plugin(d) and Plugin(e)) introduced at the end of Section \ref{sec:lowdim}, and plug them in the Bayes rule (\ref{eq:bayes2}), as done in (\ref{eq:qsc_plugin}). For comparisons, we also supply the estimated $\beta_0$, $\beta$ and $\nabla$ from the CLIPS procedure to a QDA classifier which is applied to all the observations in a testing set, followed by a majority voting scheme (labeled as QDA-MV). Lastly, we calculate the sample mean and variance of each variable in an observation set to form a new feature vector as done in \cite{miedema2012image}; then support vector machine \citep[SVM;][]{Cortes1995Support} and distance weighted discrimination \citep[DWD;][]{Marron2007Distance,wang2018another} are applied to the features to make predictions (labeled as SVM and DWD respectively). We use R library \texttt{clime} to calculate the CLIME estimates, R library \texttt{e1071} to calculate the SVM classifier, and R library \texttt{sdwd} \citep{Wang2015Sparse} to calculate the DWD classifier.

\subsection{Simulations}\label{sec:simulation}
Three scenarios are considered for simulations. In each scenario, we consider a binary setting with $N=7$ sets in a class, and $M=10$ observations from normal distribution in each set.

\begin{description}
	\item [Scenario 1] We set the  precision matrix for Class 1 to be $\Sigma_1^{-1}=(1+\sqrt{p})I_p$. For Class 2, we set $\Sigma_2^{-1}=\Sigma_1^{-1}+\tilde\nabla$, where $\tilde\nabla$ is a $p\times p$ symmetric matrix with $10$ elements randomly selected from the upper-triangular part whose values are $\zeta$ and other elements being zeros. For the mean vectors, we set $\mu_1=\Sigma_1(u,u,0,\dots,0)^T$ and $\mu_2=(0,\dots,0)^T$. Note that this makes the true value of $\beta = \Sigma_1^{-1}\mu_1 - \Sigma_2^{-1}\mu_2 = (u,u,0,\dots,0)^T$, that is, only the first two covariates have linear impacts on the discriminant function if $u\neq 0$. In this scenario, the true difference in the precision matrices has some sparse and large non-zero entries, whose magnitude is controlled by $\zeta$. Note that while the diagonals of the precision matrices are the same, the diagonals of the covariance matrices are different between the two classes.
	
	\item [Scenario 2] We set the covariance matrices for both classes to be the identity matrix, except that for Class 1 the leading 5 by 5 submatrix of $\Sigma_1$ has  its off-diagonal elements set to $\rho$. The rest of the setting is the same as in Scenario 1. In this scenario, both the difference in the covariance and the difference in the precision matrix are confined in the leading 5 by 5 submatrix, so that the majority of matrix entries are the same between the two classes. The level of difference is controlled by $\rho$: when $\rho=0$, the two classes have the same covariance matrix.
	
	\item [Scenario 3] We set the precision matrix $\Sigma_1$ for Class 1 to be a Toeplitz matrix whose first row is $(1-\rho^2)^{-1}(\rho^0,\rho^1,\rho^2,\dots,\rho^{p-1})$. The covariance for Class 2, $\Sigma_2$, is a diagonal matrix with the same diagonals as those of $\Sigma_1$. It can be shown that the precision matrix for Class 1 is a band matrix with degree 1, that is, a matrix whose nonzero entries are confined to the main diagonal and one more diagonal on both sides. Since the precision matrix for Class 2 is a diagonal matrix, the difference between the precision matrix has up to $p+2(p-1)$ nonzero entries. The magnitude of the difference is controlled by the parameter $\rho$. The rest of the setting is the same as in Scenario 1.
\end{description}

We consider different comparisons where we vary the magnitude of the difference in the precision matrices ($\zeta$ or $\rho$), the magnitude of the difference in mean vectors ($u$), or the dimensionality ($p$), when the other parameters are fixed.

\begin{description}
	\item [Comparison 1 (varying $\zeta$ or $\rho$)] We vary $\zeta$ or $\rho$ but fix $p=100$ and $u=0$, which means that the mean vectors have no discriminant power since the true value of $\beta$ is a zero vector. It shows the performance with different potentials in the covariance structure.
	\item [Comparison 2 (varying $u$)] We vary $u$ while fixing $p=100$ and $\zeta=0.55$ in Scenario 1 or $\rho=0.5$ and $0.3$ in Scenarios 2 and 3. This case illustrates the potentials of the mean difference when there is some useful discriminative power in the covariance matrices.
	\item [Comparison 3 (varying $p$)] We let $p=80, 100, 120, 140, 160$ while fixing $\zeta$ or $\rho$ in the same way as in Comparison 2 and fixing  $u=0.05$, 0.025 and 0.025 in Scenarios 1, 2 and 3 respectively.
\end{description}

\begin{figure}[h!]
	\centering
	\includegraphics[width=1\textwidth]{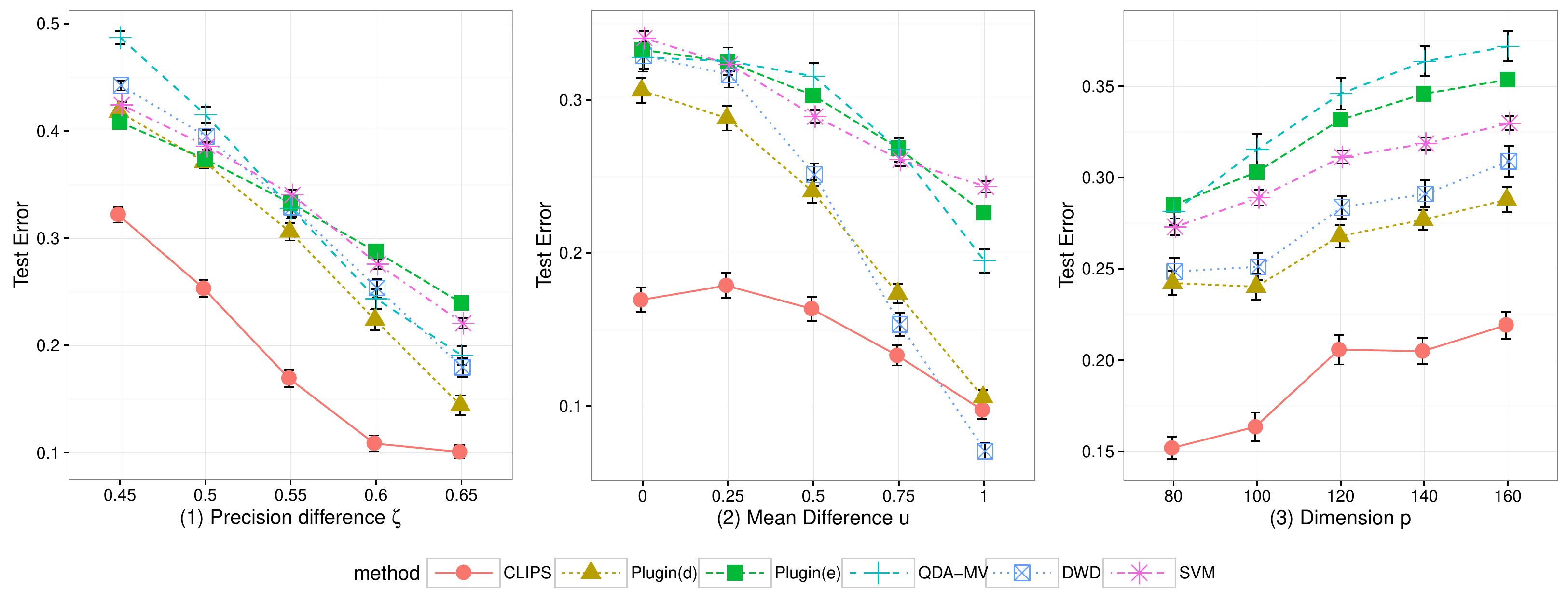}
	\caption{\footnotesize Set classification for Scenario 1. The three panels are corresponding to varying $\zeta$, varying $u$ and varying $p$ respectively. The CLIPS classifier performs very well when the effect of covariance dominates that of the mean difference.}\label{fig:fig_setting1}
\end{figure}

Figure \ref{fig:fig_setting1} shows the performance for Scenario 1. In the left panel, as $\zeta$ increases, the difference between the true precision matrices increases. The proposed CLIPS classifier performs the best among all methods under consideration. It may be surprising that the Plugin(d) method, which does not consider the off-diagonal elements in the sample covariance, can work reasonably well in this setting where the major mode of variation is in the off-diagonal of the precision matrices. However, since large values in the off-diagonal of the precision matrix can lead to large values of some diagonal entries of the covariance matrix, the good performance of Plugin(d) has some partial justification.

In the middle panel of Figure \ref{fig:fig_setting1}, the mean difference starts to increase. While every method more or less gets some improvement, the DWD method has gained the most (it is even the best performing classifier when the mean difference $u$ is as large as 1.) This may be due to the fact that the mean difference on which DWD relies, instead of the difference in the precision matrix, is sufficiently large to secure a good performance in separating sets between two classes.

\begin{figure}[t!]
	\centering
	\includegraphics[width=1\textwidth]{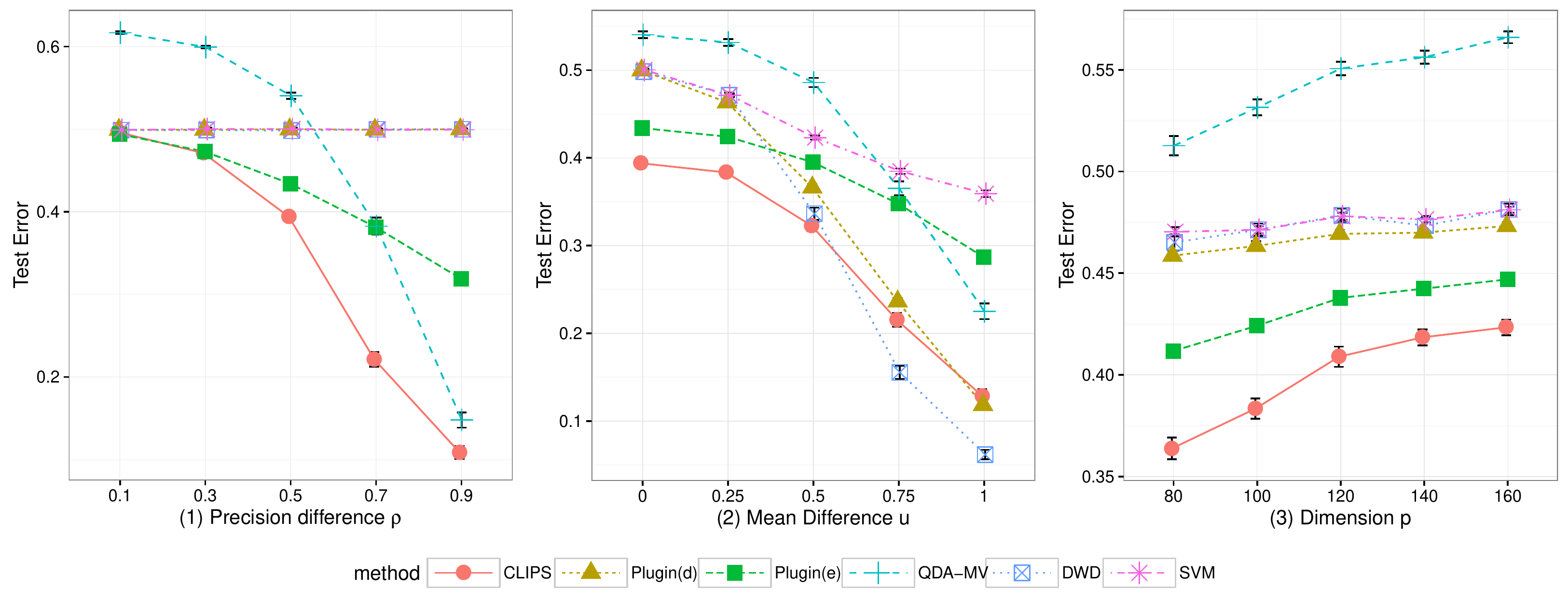}
	\caption{\footnotesize Set classification for Scenario 2. The three panels are corresponding to varying $\rho$, varying $u$ and varying $p$ respectively. The classifiers that do not engage covariance perform poorly when there is no mean difference signal.}\label{fig:fig_setting2}
\end{figure}

\begin{figure}[b!]
	\centering
	\includegraphics[width=1\textwidth]{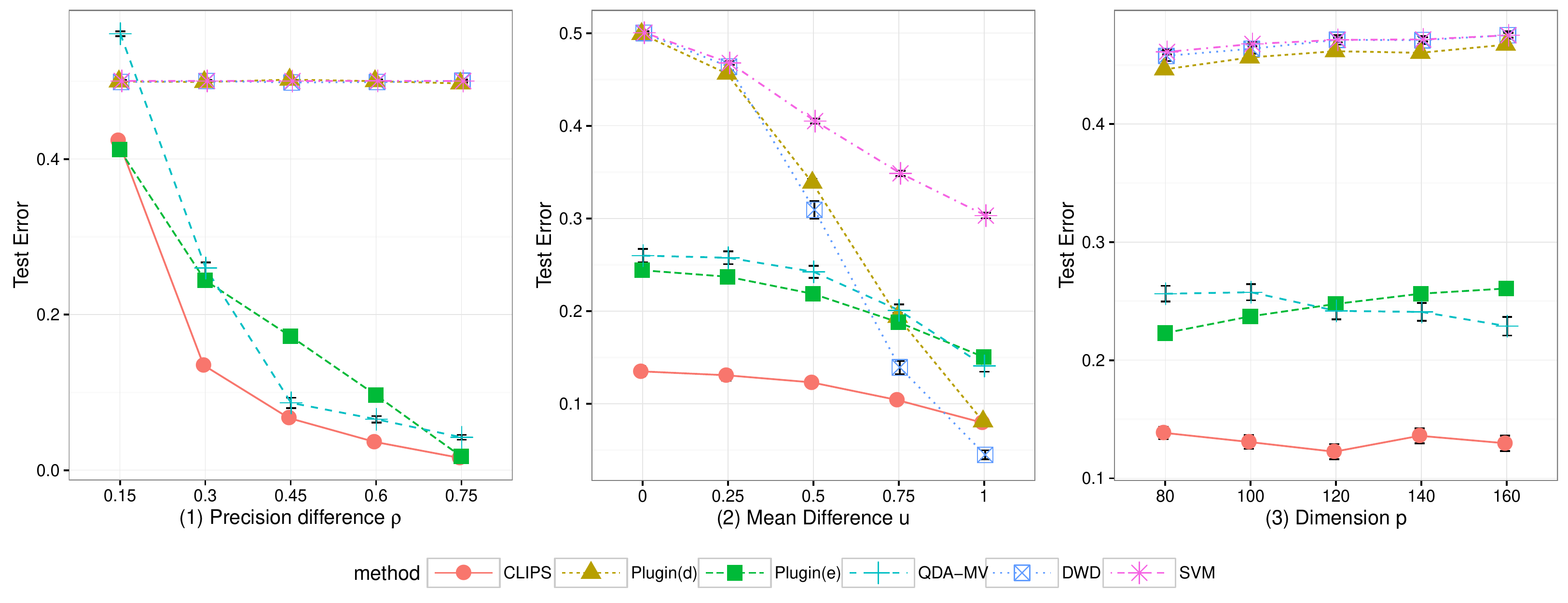}
	\caption{\footnotesize Set classification for Scenario 3. The three panels are corresponding to varying $\rho$, varying $u$ and varying $p$ respectively. As in Scenario 2, the classifiers that do not engage covariance perform poorly when there is no mean difference signal.}\label{fig:fig_setting3}
\end{figure}

Figure \ref{fig:fig_setting2} shows the results for Scenario 2. In contrast to Scenario 1, there is no difference in the diagonals of the covariances between the two classes (the precision matrices are still different). When there is no mean difference (see the left panel), it is clear that DWD, SVM and the Plugin(d) method fail for obvious reasons (note that the Plugin(d) method does not read the off-diagonal of the sample covariances and hence both classes have the same precision matrices from its viewpoint.) As a matter of fact, all these methods perform as badly as random-guess. The CLIPS classifier always performs the best in this scenario in the left panel.
%It is interesting that QDA-MV performs even worse than random-guess when $\rho\le 0.5$ but significantly improves for large values of $\rho$.
Similar to the case in Scenario 1, as the mean difference increases (see the middle panel), the DWD method starts to get some improvement.

The results for Scenario 3 (Figure \ref{fig:fig_setting3}) are similar to Scenario 2, except that, this time the advantage of two covariance-engaged set classification methods, CLIPS and Plugin(e), seems to be more obvious when the mean difference is 0 (see left panel). Moreover, the QDA-MV method also enjoys some good performance, although not as good as the CLIPS classifier.

In all three scenarios, it seems that the test classification error is linearly increasing in the dimension $p$, except for Scenario 3 in which the signal level depends on $p$ too (greater dimensions lead to greater signals.)
%%%%%%%%%%%%%%%%%%%%%%%%%%%%%%%%%%%%%%%%%
\subsection{Data Example}\label{sec:data}
One of the common procedures used to diagnose hepatoblastoma (a rare malignant liver cancer) is biopsy. A sample tissue of a tumor is removed and examined under a microscope. A tissue sample contains a number of nuclei, a subset of which is then processed to obtain segmented images of nuclei. The data we analyzed contain 5 sets of nuclei from normal liver tissues and 5 sets of nuclei from cancerous tissues. Each set contains 50 images. The data set is publicly available (http://www.andrew.cmu.edu/user/gustavor/software.html)
and was introduced in \cite{Wang2011,Wang2010}.
%The same data set was previous analyzed by \cite{Jung2014}.

\begin{table}[!b]
	\centering
	\begin{tabular}{c|cc}
		Method & number of misclassified sets  &   standard error \\\hline
		CLIPS &       0.01/10 & 0.0104 \\
		Plugin(d) &       0.74/10 & 0.0450 \\
		Plugin(e) &       0.97/10 & 0.0178 \\
		QDA-MV &       0.08/10 & 0.0284 \\
		DWD &       3.24/10 & 0.1164 \\
		SVM &       3.13/10 & 0.1130
	\end{tabular}\caption{\footnotesize Classification performance for the liver cell nucleus image data.}\label{tab:liver}
\end{table}

We tested the performance of the proposed method on the liver cell nuclei image data set. First, the dimension was reduced from 36,864 to 30 using principal component analysis. Then, among the 50 images of each set, 16 images are retained as training set, 16 are tuning set and another 16 are test set. In other words, for each of the training, tuning, and testing data sets, there are 10 sets of images, five from each class, with 16 images in each set.

\begin{figure}[!b]
	\centering
	\includegraphics[width=0.8\textwidth]{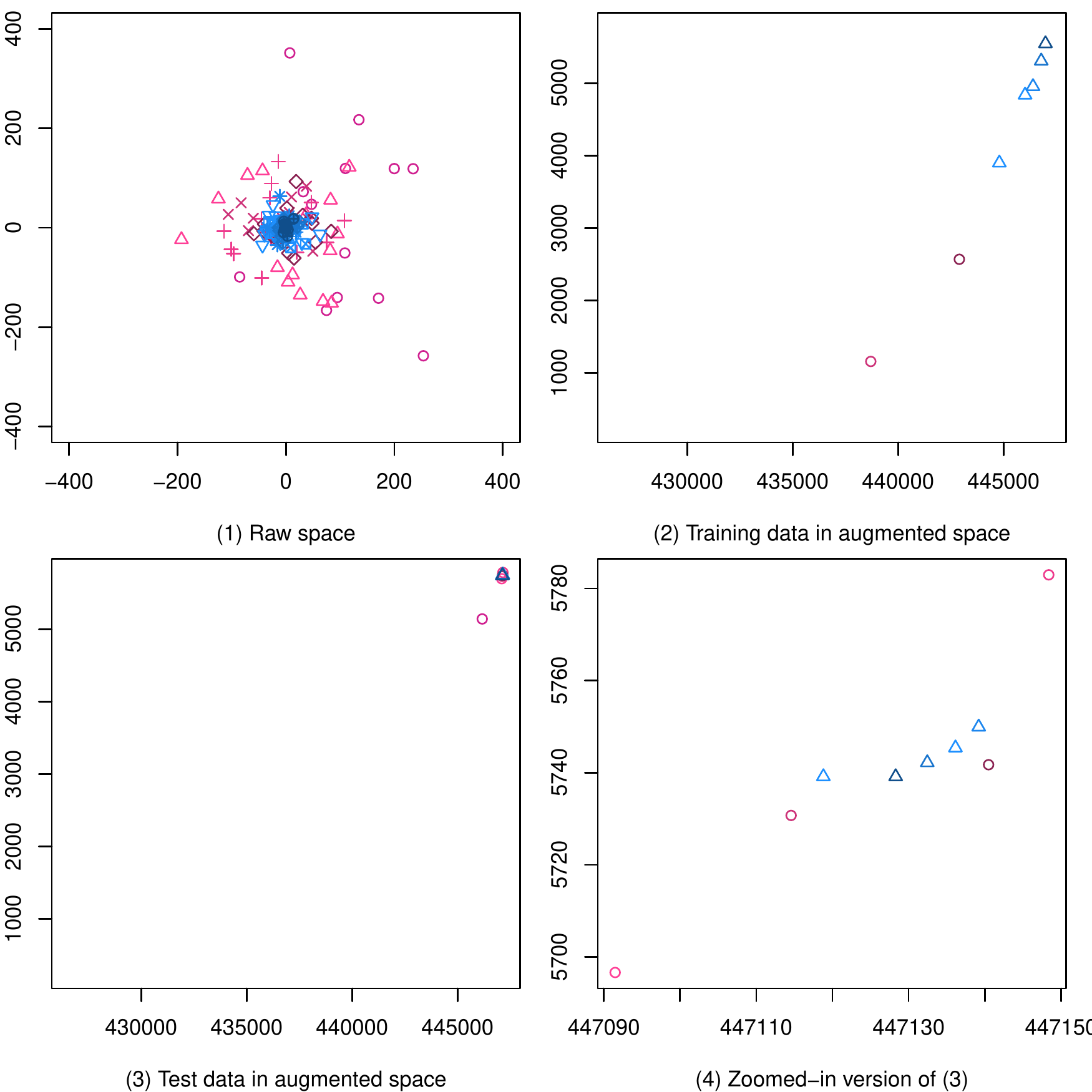}
	\caption{\footnotesize PCA scatter plots for the liver cell nucleus image data. Both classes are shown in different colors. (1): the elementary observations in the raw space; different sets are shown in different symbols.  (2) and (3): the augmented space seen by the DWD and SVM methods. (4) is a zoomed-in version of (3). It is shown that traditional multivariate methods have a fundamental difficulty for this data set.}\label{fig:liver_scat}
\end{figure}

Table \ref{tab:liver} summarizes the comparison between the methods under consideration. All three covariance-engaged set classifiers (CLIPS, Plugin(d) and Plugin(e)), along with the QDA-MV method, perform better than methods which do not take the covariance matrices much into account, such as DWD and SVM (note that they do look into the diagonal of the covariance matrix.)

To get some insights to the reason that covariance-engaged set classifiers work and traditional methods fail, we visualize the data set in Figure \ref{fig:liver_scat}. Subfigure (1) shows the scatter plot of the first two principal components of all the elementary observations (ignoring the set memberships) in the data sets, in which different colors (blue versus violet) depict the two different classes. Observations in the same set are shown in the same symbol. The first strong impression is that there is no mean difference between the two classes on the observation level. In contrast, it seems that it is the second moment such as the variance that distinguishes the two classes.

One may argue that DWD and SVM should theoretically work here because they work on the augmented space where the mean and variance of each variable are calculated for each observation set, leading to a $2p$-dimensional feature vector for each set. However, Subfigures (2)--(4) invalidate this argument. We plot the augmented training data in the space formed by the first two principal components (Subfigure (2)). The augmented test data are shown in the same space in Subfigure (3) with a zoomed-in version in Subfigure (4). Note that the scales for Subfigures (2) and (3) are the same. These figures show that there are more than just the marginal mean and variance that are useful here, and our covariance-engaged set classification methods have used the information in the right way.

%%%%%%%%%%%%%%%%%%%%%%%%%%%%%%%%%%%%%%%%%%%%%%%%%%%%%%%%%%%%%%%%%%%%%%%%%%%%%%%%%%%%%%%%%%%%%%%%%%%%%%%%%%%%%%%%%%%%%%%%%%%%
\vskip 14pt
\noindent {\large\bf Supplementary Materials}

The online supplementary materials contain additional theoretical arguments and proofs of all results.
\par
%%%%%%%%%%%%%%%%%%%%%%%%%%%%%%%%%%%%%%%%%%%%%%%%%%%%%%%%%%%%%%%%%%%%%%%%%%%%%%%%%%%%%%%%%%%%%%%%%%%%%%%%%%%%%%%%%%%%%%%%%%%%
\vskip 14pt
\noindent {\large\bf Acknowledgments}

This work was supported by the \emph{National Research Foundation of Korea} (No. 2019R1A2C2002256) and a collaboration grant from \textit{Simons Foundation} (award number 246649).
\par

%%%%%%%%%%%%%%%%%%%%%%%%%%%%%%%%%%%%%%%%%%%%%%%%%%%%%%%%%%%%%%%%%%%%%%%%%%%%%%%%%%%%%%%%%%%%%%%%%%%%%%%%%%%%%%%%%%%%%%%%%%%%
\markboth{\hfill{\footnotesize\rm Sungkyu Jung} \hfill}
{\hfill {\footnotesize\rm Bias in principal component scores} \hfill}

%\iffalse
\bibhang=1.7pc
\bibsep=2pt
\fontsize{9}{14pt plus.8pt minus .6pt}\selectfont
\renewcommand\bibname{\large \bf References}
%\begin{thebibliography}{11}
\expandafter\ifx\csname
natexlab\endcsname\relax\def\natexlab#1{#1}\fi
\expandafter\ifx\csname url\endcsname\relax
  \def\url#1{\texttt{#1}}\fi
\expandafter\ifx\csname urlprefix\endcsname\relax\def\urlprefix{URL}\fi
%\fi

\lhead[\footnotesize\thepage\fancyplain{}\leftmark]{}\rhead[]{\fancyplain{}\rightmark\footnotesize{} }%Put this line in Page 2
%%%%%%%%%%%%%%%%%%%%%%%%%%%%%%%%%%%%%%%%%%%%%%%%%%%%%%%%%%

%\bibliographystyle{biometrika}
\bibliographystyle{apalike}
\bibliography{qsc_bib}

%%%%%%%%%%%%%%%%%%%%%%%%%%%%%%%%%%%%%%%%%%%%%%%%%%%%%%%%%%%%%%%%%%%%%%%%%%%%%%%%%%%%%%%%%%%%%%%%%%%%%%%%%%%%%%%%%%%%%%%%%%%%
\vskip .65cm
\noindent
Zhao Ren\\
\noindent
Department of Statistics, University of Pittsburgh, Pittsburgh, PA 15260, USA
\vskip 2pt
\noindent
E-mail:zren@pitt.edu
\vskip 2pt

\noindent
Sungkyu Jung \\
\noindent
Department of Statistics, Seoul National University, Gwanak-gu, Seoul 08826, Korea
\vskip 2pt
\noindent
E-mail: sungkyu@snu.ac.kr
\vskip 2pt

\noindent
Xingye Qiao \\
\noindent
Department of Mathematical Sciences,        Binghamton University,        State University of New York,        Binghamton, NY, 13902 USA
\vskip 2pt
\noindent
E-mail: qiao@math.binghamton.edu
 \vskip .3cm
%\centerline{(Received ???? 20??; accepted ???? 20??)}\par

\newpage
\includepdf[pages={1-48}]{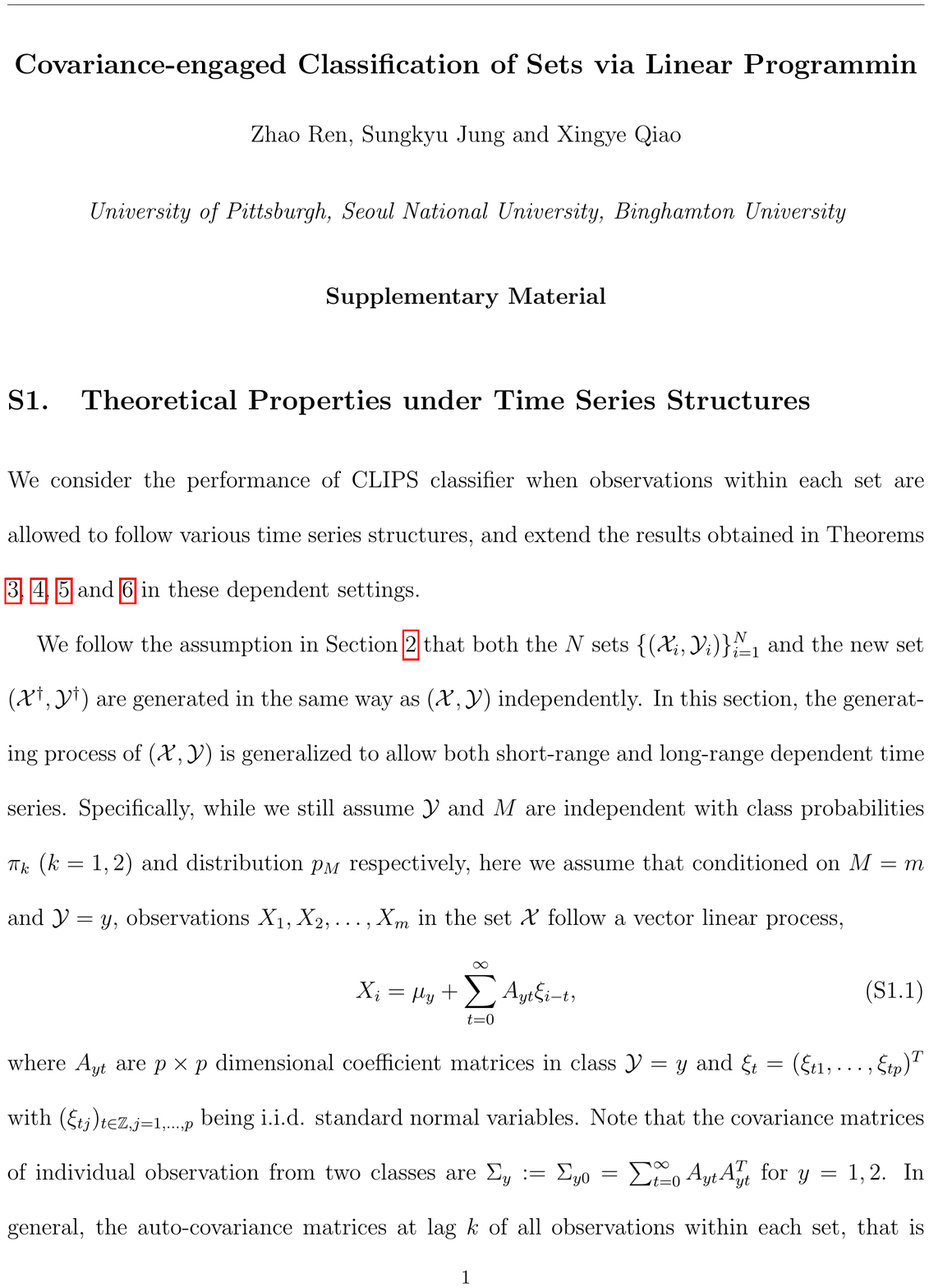}

\end{document}